\documentclass{article} 
\usepackage{iclr2025_conference,times}

\usepackage{amsmath,amsfonts,bm}

\def\eqref#1{equation~\ref{#1}}

\def\1{\bm{1}}

\DeclareMathAlphabet{\mathsfit}{\encodingdefault}{\sfdefault}{m}{sl}
\SetMathAlphabet{\mathsfit}{bold}{\encodingdefault}{\sfdefault}{bx}{n}

\usepackage{hyperref}
\usepackage{url}
\usepackage{multicol}
\usepackage{url}
\usepackage{booktabs}
\usepackage{multirow}
\usepackage{makecell}
\usepackage{arydshln}
\usepackage{xcolor}
\usepackage{pdfpages}
\usepackage{CJKutf8}
\usepackage{xspace}
\usepackage[utf8]{inputenc}
\usepackage{longtable}
\usepackage{array}
\usepackage{caption}
\usepackage{float}
\usepackage{tcolorbox}
\usepackage{CJKutf8}
\usepackage{siunitx}
\newcommand{\model}{\textit{Steel-LLM}\xspace}
\usepackage{amsmath}  
\usepackage{amssymb}  
\usepackage{graphicx} 
\usepackage{subcaption} 
\usepackage{wrapfig}    
\usepackage{wasysym}
\usepackage{authblk}

\newcommand{\authorinfo}[3]{
  \textbf{#1} \\
  \textit{#2} \\
  \texttt{#3}
}

\title{\model: From Scratch to Open Source \\– A Personal Journey in Building a Chinese-Centric LLM}

\begin{document}
\maketitle


\begin{center}
  \begin{minipage}{0.8\textwidth}
    \centering
    \begin{multicols}{2} 
      \authorinfo{Qingshui Gu}{Tsinghua University}{gqs19@tsinghua.org.cn} \\
      \vspace{0.5cm} 
      \authorinfo{Tianyu Zheng}{Beijing University of Posts \\ and Telecommunications}{zhengtianyu@bupt.edu.cn} \\
      
      \columnbreak 
      
      \authorinfo{Shu Li}{Tsinghua University}{lishu14@tsinghua.org.cn} \\
      
      \vspace{0.5cm} 
      \authorinfo{Zhaoxiang Zhang}{Institute of Automation, \\ Chinese Academy of Sciences}{zhaoxiang.zhang@ia.ac.cn}
    \end{multicols}
  \end{minipage}
\end{center}
\vspace{-0ex}
\begin{center}
     \url{https://github.com/zhanshijinwat/Steel-LLM}\\
\end{center}
\vspace{1ex}
\iclrfinalcopy

%


\begin{abstract}
\model is a Chinese-centric language model developed from scratch with the goal of creating a high-quality open-source model despite limited computational resources. Launched in March 2024, the project aimed to train a 1-billion-parameter model on a large-scale dataset, prioritizing transparency and the sharing of practical insights to assist others in the community. The training process primarily focused on Chinese data, with a small proportion of English data included, addressing gaps in existing open-source LLMs by providing a more detailed and practical account of the model-building journey. \model has demonstrated competitive performance on benchmarks such as CEVAL and CMMLU, outperforming early models from larger institutions. This paper provides a comprehensive summary of the project's key contributions, including data collection, model design, training methodologies, and the challenges encountered along the way, offering a valuable resource for researchers and practitioners looking to develop their own LLMs.
The model checkpoints and the training script are available at \url{https://github.com/zhanshijinwat/Steel-LLM}.
\end{abstract}

\section{Introduction}

The rapid advancements in open-source large language models (LLMs) have led to significant achievements in natural language processing (NLP), enabling applications ranging from conversational agents to code generation. However, despite these advances, challenges remain in the transparency, accessibility, and resource efficiency of LLM development. Many prominent LLMs, such as the Qwen series~\citep{bai2023qwentechnicalreport,yang2024qwen2}, Llama~\citep{touvron2023llama2openfoundation}, and Deepseek~\citep{deepseekai2024deepseek}, provide only the final model weights while withholding essential details such as training data, code, and intermediate checkpoints. This limited openness creates obstacles to reproducibility and prevents the broader research community from building upon these models effectively.

Several initiatives, such as LLM360's Amber~\citep{liu2023llm360,tan2024llm360}, M-A-P's Neo~\citep{zhang2024map}, and AI2's OLMO series~\citep{OLMo,olmo20242,muennighoff2024olmoe}, have addressed these limitations by releasing complete training pipelines, datasets, and intermediate checkpoints. While these contributions are invaluable, their development typically requires extensive computational resources, making them inaccessible to smaller research teams or individual practitioners. This creates a significant gap for fully transparent and resource-efficient LLMs tailored for smaller-scale research efforts, particularly in non-English languages such as Chinese.

This paper introduces \model, a small, fully open-source, Chinese-centric LLM developed with limited computational resources. Launched in March 2024, the project aimed to train a 1-billion-parameter model on a large-scale dataset, prioritizing transparency and the sharing of practical insights to assist others in the community. Unlike many large-scale projects, \model demonstrates that high-quality LLMs can be developed efficiently while maintaining transparency. This work focuses on the Chinese language, with a small proportion of English data, addressing gaps in existing open-source LLMs by providing a more detailed and practical account of the model-building journey.

\model achieves competitive performance on benchmarks such as CEVAL and CMMLU, outperforming early models from larger institutions. The model architecture incorporates innovative features such as Soft Mixture of Experts (Soft MoE) and an enhanced Feed-Forward Network, optimizing performance within resource constraints. Our training framework, modified from TinyLlama~\citep{zhang2024tinyllamaopensourcesmalllanguage}, includes several optimizations for efficiency and usability, such as improved model loading, restoration of training progress, and the ability to append data during the training process.

The contributions of this work are as follows:

\begin{itemize}
    \item \textbf{Resource-Efficient Model Development}: We present a 1-billion-parameter model trained primarily on Chinese data, with a small proportion of English, addressing the need for more diverse language representation in open-source LLMs. By leveraging limited computational resources (8 GPUs), we demonstrate that high-quality LLMs can be developed without access to large-scale infrastructure.
    
    \item \textbf{Complete Transparency}: We offer complete transparency in our development process, including the release of our training pipeline, dataset, model architecture, and intermediate checkpoints. This facilitates reproducibility and allows for further research by the broader community.
    
    \item \textbf{Practical Guidance for Small-Scale Research}: We provide detailed insights into our model architecture, training framework, and data preparation process, offering practical guidance for researchers and practitioners with limited resources. This includes optimizations for training efficiency, such as mixed-precision training, FlashAttention, and operator fusion.
    
    \item \textbf{Benchmark Performance}: \model achieves competitive performance on Chinese benchmarks such as CEVAL and CMMLU, outperforming some early models from larger institutions. This demonstrates the effectiveness of our approach in developing a high-quality Chinese-centric LLM with limited resources.
\end{itemize}

By prioritizing resource efficiency, openness, and practical applicability, this work contributes to the broader LLM research community, offering a valuable resource for replicating or extending similar efforts with fewer computational constraints. All model checkpoints, training scripts, and related resources are fully open-sourced, further promoting transparency and collaboration in the field of LLM development.

\section{Related Works}
Recent developments in open source LLM have varied widely in terms of transparency and accessibility. Many models, such as Qwen~\citep{bai2023qwentechnicalreport,yang2024qwen2}, Llama~\citep{touvron2023llama2openfoundation}, Deepseek~\citep{deepseekai2024deepseek}, Gemma ~\citep{gemmateam2024gemmaopenmodelsbased}, InternLM ~\citep{cai2024internlm2}, Mixtral ~\citep{jiang2023mistral}, Yi ~\citep{ai2024yiopenfoundationmodels}, GLM ~\citep{glm2024chatglm}, have been released with only the final model checkpoints and weights, while withholding crucial details such as training data, codes, and intermediate checkpoints. This limited transparency hinders reproducibility and makes it difficult for the broader research community to fully understand or build upon these models.

In response to these challenges, several initiatives have adopted a more open approach by releasing complete training pipelines, datasets, and intermediate checkpoints. Notable examples include LLM360’s Amber~\citep{liu2023llm360, tan2024llm360}, M-A-P’s MAP-Neo~\citep{zhang2024map}, and AI2’s OLMO series~\citep{OLMo, olmo20242, muennighoff2024olmoe}, all of which offer comprehensive resources, including training code, model weights, and intermediate checkpoints. While these contributions are invaluable to the field, the large-scale nature of these models necessitates significant computational resources, which may render them inaccessible to smaller research teams or individual practitioners. Consequently, there remains a gap for open-source LLMs that are both fully transparent and feasible for smaller teams to develop and deploy.

This paper addresses this need by presenting a small, fully open-source, Chinese-centric LLM. Designed to minimize resource requirements, our model offers a complete end-to-end solution for researchers with limited computational access. By sharing the full training pipeline, dataset, and model architecture, we aim to make the development of high-quality LLMs more accessible. This work emphasizes transparency and practical guidance, providing a valuable resource for those seeking to replicate or extend the model with fewer computational constraints.

\section{Architecture}
\begin{figure}[H]
    \centering
    \includegraphics[width=1\textwidth]{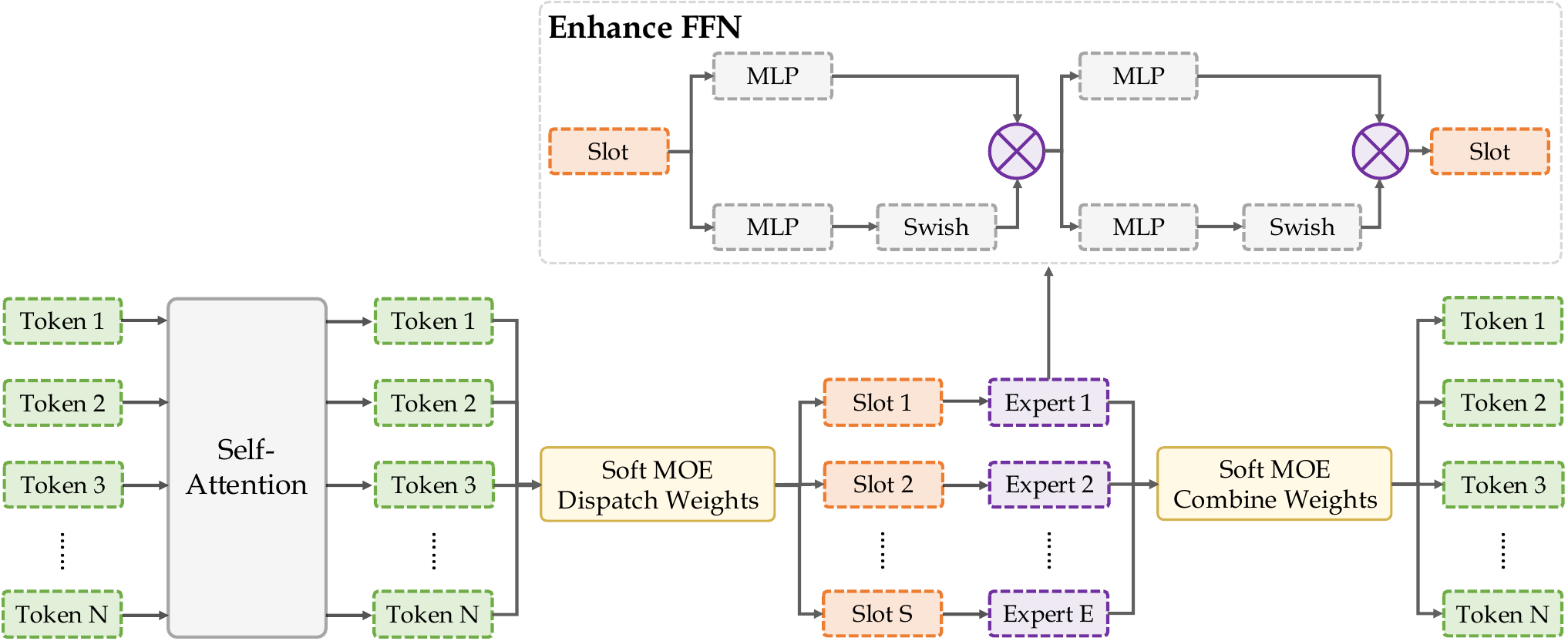}
    \caption{The architecture of \model Transformer block}
    \label{fig:steelllm_arch}
\end{figure}

The model structure of \model is adapted from Qwen(\cite{bai2023qwentechnicalreport}).A Transformer block can be roughly divided into self-attention and Feed-Forward Network (FFN). An efficient implementation of self-attention is Flash Attention(\cite{dao2022flashattentionfastmemoryefficientexact}), which is widely utilized. Flash Attention not only improves the efficiency of model training and inference but also saves GPU memory. \model reuses Flash Attention and only makes improvements to the FFN layer. In the FFN layer, we adopts Soft Mixture of Experts (Soft MoE, \cite{puigcerver2024sparsesoftmixturesexperts}) and  enhances the second layer of the MLP. The architecture of \model's Transformer block is illustrated in \autoref{fig:steelllm_arch}.

\subsection{Soft MoE}

Mixture of Experts (MoE)~\citep{6797059} was first proposed in 1991 and is widely used in the field of recommendation systems ~\citep{ma2018modeling}. In the architecture of large language models, sparse MoE is commonly employed, such as in deepSeekMoE(\cite{dai2024deepseekmoeultimateexpertspecialization}), Qwen-MoE(\cite{qwen_moe}), etc. A typical practice to construct an MoE language model usually replaces FFNs in a Transformer with MoE layers(\cite{lepikhin2021gshard};\cite{9835248};\cite{fedus2022switchtransformersscalingtrillion}). An MoE layer consists of multiple experts, and each expert is a FFN. A gating network calculates scores to assign each token to a small subset of experts. Under the condition of the same parameter scale, the sparse MoE model activates fewer parameters than a dense model, thus reducing the number of FLOPs.

\model is trained using only 8 GPUs with limited GPU memory. For a model featuring sparse MoE structure, all parameters must be loaded into the GPU memory. Owing to the sparse nature of expert selection, in contrast to a dense model, the FFN parameters of the sparse MoE model cannot be fully trained. \model is a small-scale language model with 1 billion parameters, leading to a relatively low computational load. Meanwhile, our objective is to fully train each parameter and leverage the advantages of the MoE structure to enhance model performance. Consequently, we ultimately opt for the soft MoE(\cite{puigcerver2024sparsesoftmixturesexperts}) structure. Soft MoE is fully differentiable. Therefore, there is no need to consider problems such as expert imbalance that exist in sparse MoE.

We denote the input tokens for a single sequence as \(\mathbf{X} \in \mathbb{R}^{m \times d}\), where \(m\) represents the number of tokens and \(d\) denotes their dimensions. In the soft MoE layer, each expert processes  \(p\) slots, and each slot has a corresponding  d-dimensional learnable parameter matrix \(\Phi\in\mathbb{R}^{d\times(n\cdot p)}\). The number of slots is a crucial hyperparameter for adjusting the time complexity of the soft MoE layer. The input slots \(\tilde{\mathbf{X}}\in\mathbb{R}^{(n\cdot p)\times d}\) are obtained through convex combinations of all the \(m\) input tokens, which can be computed as follows:

\[\mathbf{D}_{ij}=\frac{\exp((\mathbf{X}\boldsymbol{\Phi})_{ij})}{\sum_{i^{\prime}=1}^m\exp((\mathbf{X}\boldsymbol{\Phi})_{i^{\prime}j})},\quad\tilde{\mathbf{X}}=\mathbf{D}^\top\mathbf{X}\]

We denote \(D\) as the dispatch weight matrix. It is obtained by applying the softmax function column-wise to the matrix  \(\mathbf{X}\mathbf{\Phi}\). Then, the corresponding expert function is applied to each slot
(i.e., on rows of \(\tilde{\mathbf{X}}\)) to obtain the output slots \(\tilde{\mathbf{Y}}_i=f_{\lfloor i/p\rfloor}(\tilde{\mathbf{X}}_i)\). The combination process is then carried out as follows:
\[\mathbf{C}_{ij}=\frac{\exp((\mathbf{X}\boldsymbol{\Phi})_{ij})}{\sum_{j^{\prime}=1}^{n\cdot p}\exp((\mathbf{X}\boldsymbol{\Phi})_{ij^{\prime}})},\quad\mathrm{~}\mathbf{Y}=\mathbf{C}\tilde{\mathbf{Y}}\]
We refer to \(C\) as the combine weights, which is the result of applying the softmax function row-wise to the matrix \(\mathbf{X}\mathbf{\Phi}\). The output tokens \(Y\) are computed as a convex combination of all \((n\cdot p)\) output slots.

\subsection{Enhanced Feed-Forward Network}

Since \model employs the soft MoE approach, within this framework, the Feed-Forward Network (FFN) effectively represents an expert. In a vanilla Transformer, The FFN comprises two layers of Multi-Layer Perceptrons (MLPs). In the architecture of large language models, a  prevalent strategy is to enhance the first layer of the MLP using the SwiGLU(\cite{shazeer2020gluvariantsimprovetransformer}) activation function(\cite{bai2023qwentechnicalreport};\cite{touvron2023llamaopenefficientfoundation};\cite{deepseekai2024deepseekv2strongeconomicalefficient}). The SwiGLU activation function enhances the model's non-linear representational capabilities, thereby improving its performance. Additionally, we extended the application of the SwiGLU activation function to the second layer of the MLP within the FFN.

Regarding other architectural elements, \model adopts the modifications made by Qwen(\cite{bai2023qwentechnicalreport}) to the transformer block. These modifications are widely used in open-source models such as LLama, Mixtral, and Deepseek.

\begin{wraptable}{r}{0.32\textwidth}
\vspace{-3pt}
\centering
\begin{tabular}{@{}lc@{}}
\toprule
Parameters & Value \\
\midrule
Layers & 18 \\

Heads & 32 \\
KV heads & 32 \\
Num\_experts & 6 \\
Slots\_per\_expert & 1 \\
Hidden size & 1,792 \\
Intermediate size & 1,792 \\
Vocab size & 151,936 \\
\bottomrule
\end{tabular}
\caption{Key model parameters.}
\vspace{-20pt}
\label{tab:modelparameter}
\end{wraptable}

 \textbullet\ \textbf{Positional embedding}.We intend to use the Rotary Position Embedding (RoPE)(\cite{su2023roformerenhancedtransformerrotary}) for \model. RoPE is a relative position encoding technique. Its core idea is to encode absolute position information into a rotation matrix, thereby representing the relative position relationships among tokens. During the training process, we adopt a global training precision of BF16, while for RoPE, we employ local FP32 precision.

 \textbullet\ \textbf{Bias}. Most layers of \model have no bias, except for the QKV layer(\cite{chowdhery2022palmscalinglanguagemodeling}).

 \textbullet\ \textbf{Pre-Norm \& RMSNorm}. Pre-normalization improves training stability compared to
post-normalization. We normalize the inputs of self-attention layers and FFN layers with RMSNorm(\cite{zhang2019rootmeansquarelayer}).

The hyperparameters of \model are presented in \autoref{tab:modelparameter}.  We employ the Qwen1.5 tokenizer (\cite{qwen1.5}), which utilizes byte pair encoding (BPE).

\section{Training Framework}
\model is trained using only 8 NVIDIA A100/H800 GPUs, so it is essential to maximize the training speed. Moreover, considering the potential reuse of our training code by others, we have made some usability optimizations.

Our pretraining framework, which is modified from TinyLlama(\cite{zhang2024tinyllamaopensourcesmalllanguage}), has undergone the following optimizations.

\textbullet\ \textbf{Model loading}. In the open-source community, the file format of LLM usually follows the standards of the Transformers(\cite{wolf-etal-2020-transformers}) library and it can be easily loaded through the AutoModelForCausalLM.from\_pretrained function. To enhance usability and facilitate model loading for different architectures, our framework supports the Transformer library's model loading method. The original framework, by contrast, was designed for training standard Llama architecture.

\textbullet\ \textbf{Training Progress Restoration}. In the training process, we are required to save not only the checkpoints of both the model and the optimizer but also the training progress of the data. The original framework's method of saving the training step is the simplest for recording data progress, provided that there is no data change (no addition or deletion). This limitation restricts the flexibility of data organization during the training process. We choose to serialize the entire data management class PackedDatasetIterator using the pickle library and then save it. This includes the file names of all training data, the index of each data piece, etc.

\textbullet\ \textbf{Appending Data During the Training Process}.
Training LLMs typically spans dozens of days. During this training, appending new data is a common practice. The simplest implementation approach is to train the newly added data at the end. Nevertheless, to safeguard against the significant distributional disparities between new and old data that could impact the model's training performance, we have devised a method for re-shuffling the indices of both the newly appended data and the hitherto untrained data from the old dataset. Additionally, to avert the inadvertent repeated addition of data files to the training process, \model has incorporated a function that utilizes a hash value, specifically the MD5(\cite{RFC1321}) hash algorithm, to detect data content duplication.

To improve training speed and conserve GPU memory, during the training process, we utilized techniques such as bfloat16 mixed-precision(\cite{kalamkar2019studybfloat16deeplearning}), Fully Sharded Data Parallel (FSDP, \cite{zhao2023pytorchfsdpexperiencesscaling}), and FlashAttention(\cite{dao2022flashattentionfastmemoryefficientexact}). Operator fusion represents an additional training optimization approach. By integrating multiple computational steps, it reduces intermediate activations and memory access, and can be implemented via CUDA or Triton. Specifically, during training, we adopted the CUDA-based version of Rotary Position Embedding (RoPE,\cite{su2023roformerenhancedtransformerrotary}) and the Triton-based(\cite{10.1145/3315508.3329973}) cross entropy loss function. We then conducted an ablation study using a micro-batch size of 8 and a 1.8-billion-parameter model on NVIDIA A100 GPU to analyze the impact of these techniques on training efficiency and GPU memory usage, as presented in Table \ref{table:table2}. Employing all training acceleration techniques, the training speed can be enhanced by approximately 50\%.
\begin{table}[!ht]
    \centering
    \begin{tabular}{@{}lcccccccc@{}}
    \toprule
        ~ & Exp 1 & Exp 2 & Exp 3 & Exp 4 & Exp 5 & Exp 6 & Exp 7 & Exp 8 \\ 
        \midrule
        FlashAttention & $\checkmark$ & $\checkmark$ & $\times$ & $\checkmark$ & $\checkmark$ & $\checkmark$ & $\checkmark$ & $\times$ \\ 
        SelfAttention(PyTorch) & $\times$ & $\times$ & $\checkmark$ & $\times$ & $\times$ & $\times$ & $\times$ & $\checkmark$ \\ 
        RoPE(CUDA) & $\checkmark$ & $\times$ & $\checkmark$ & $\checkmark$ & $\checkmark$ & $\checkmark$ & $\checkmark$ & $\times$ \\
        RoPE(PyTorch) & $\times$ & $\checkmark$ & $\times$ & $\times$ & $\times$ & $\times$ & $\times$ & $\checkmark$ \\ 
        RMSNorm(CUDA) & $\times$ & $\times$ & $\times$ & $\times$ & $\checkmark$ & $\times$ & $\checkmark$ & $\times$ \\
        RMSNorm(PyTorch) & $\checkmark$ & $\checkmark$ & $\checkmark$ & $\checkmark$ & $\times$ & $\checkmark$ & $\times$ & $\checkmark$ \\ 
        Loss Function(Triton) & $\checkmark$ & $\checkmark$ & $\checkmark$ & $\checkmark$ & $\checkmark$ & $\times$ & $\checkmark$ & $\times$ \\ 
        Loss Function(PyTorch) & $\times$ & $\times$ & $\times$ & $\times$ & $\times$ & $\checkmark$ & $\times$ & $\checkmark$ \\
        FSDP & $\checkmark$ & $\checkmark$ & $\checkmark$ & $\times$ & $\checkmark$ & $\checkmark$ & $\times$ & $\checkmark$ \\ 
        FSDP(no share param) & $\times$ & $\times$ & $\times$ & $\checkmark$ & $\times$ & $\times$ & $\checkmark$ & $\times$ \\
        Speed(tokens/s/gpu) & 13400 & 12500 & 10600 & 13800 & 14600 & 13000 & 15000 & 10500 \\ 
        GPU Memory(GB) & 65 & 65 & 69 & 69 & 61 & 75 & 66 & 75 \\ 
    \bottomrule
    \end{tabular}
\caption{Comparison of different training configurations.}
\label{table:table2}
\end{table}

\section{Pretraining}
The pretraining corpus employed in our study predominantly consists of Chinese texts and is entirely derived from open-source datasets. It includes prominent datasets such as SkyPile-150B ~\citep{wei2023skywork}, Wanjuan1.0~\citep{he2023wanjuan}, Wikipedia-cn, as well as diverse chat data from multiple sources and Starcode. Further details of these datasets are provided in \autoref{App:pretrain_data}.

\begin{wrapfigure}{lr}{0.445\textwidth}
    \centering
    \vspace{-7pt}
    \includegraphics[width=0.42\textwidth]{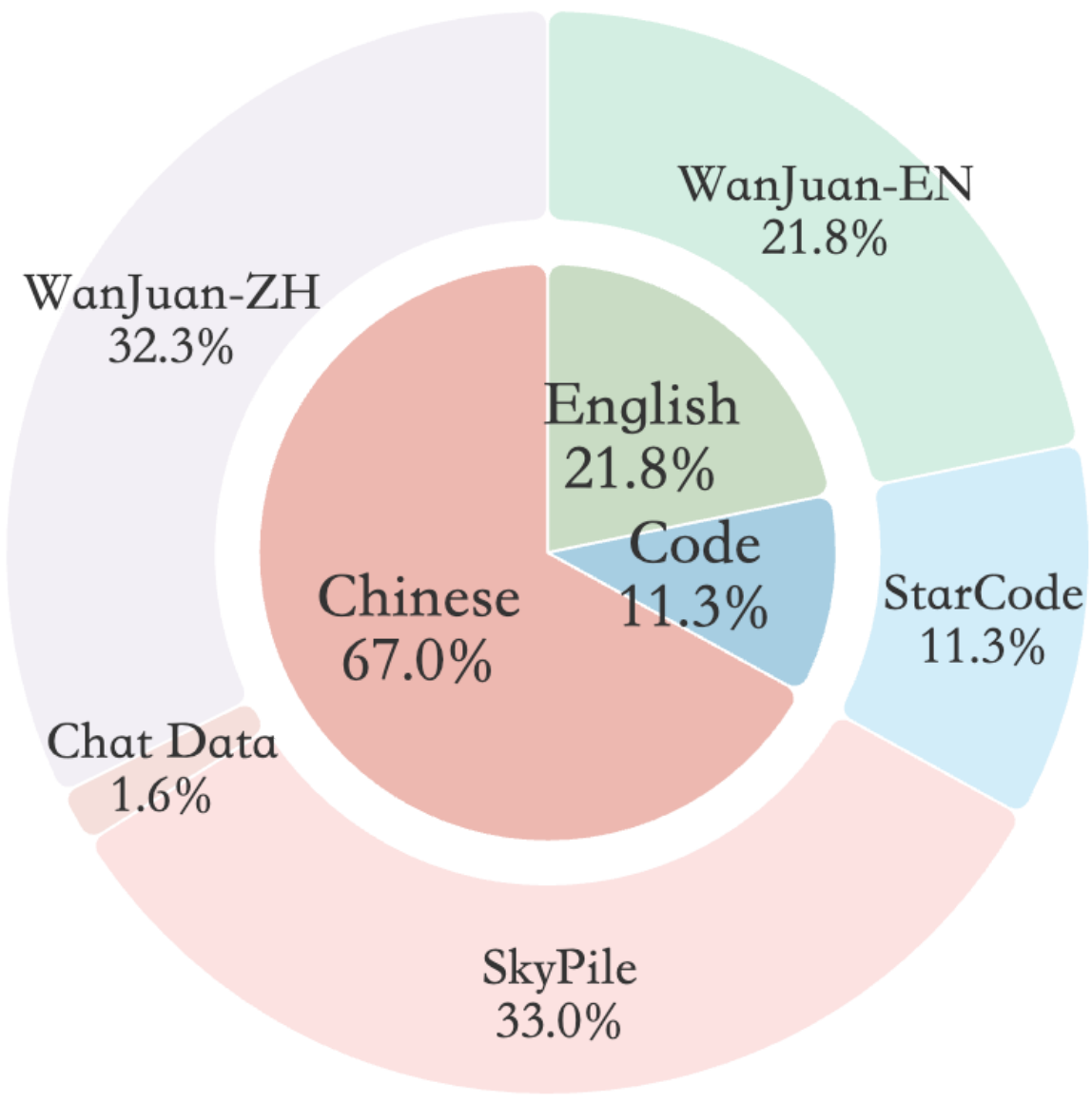}
    \caption{Pretraining Data Distribution}
    \label{fig:data_dis}
    \vspace{-10pt}
\end{wrapfigure}%

To ensure consistency across the data, we initially standardized all datasets into a uniform format. Subsequently, we utilized Alibaba's open-source tool, Data-Juicer, to meticulously filter and transcribe both the text and code data. We utilized a total of 21 text processing operators, as described in \autoref{App:operators}, and 13 code processing operators, as listed in \autoref{App:code_processing}. 
The final step involved employing the tokenizer from Qwen1.5. This tokenizer was instrumental in converting the entire corpus into token-ids, which were then merged and segmented into manageable chunks. The distribution of the pretraining data across different domains is illustrated in \autoref{fig:data_dis}. Despite these efforts, our data preprocessing workflow still exhibits certain limitations, particularly the lack of balance in the proportion of texts from different domains within the pre-training corpus.

Over a span of 30 days, \model was trained through 1.07 million steps. The initial 200,000 steps utilized NVIDIA 8 A100-80G GPUs for training, while the remaining steps employed 8 H800-80G GPUs. The loss curve is shown in \autoref{App:loss_curve}. We set the maximum sequence length to 2048, the batch size to 8, and the number of gradient accumulation steps to 8. Consequently, \model was trained on approximately one trillion tokens.  We employ the AdamW optimizer (\cite{kingma2017adammethodstochasticoptimization}) with hyper-parameters set to $\beta_1$= 0.9, $\beta_2$= 0.95, and weight\_decay = 0.05. The maximum learning rate is set to $3\times10^{-4}$, and the gradient clipping norm is set to 1.0. We employed a cosine-annealing learning rate schedule with 2,000 warmup steps, such that the final learning rate is 0.

\section{Post training}
\subsection{Supervised Finetuning}
There has been substantial discussion and some debate among researchers regarding the volume of datasets required for supervised fine-tuning. For instance, \cite{zhou2023limaalignment} advocates for selecting a few hundred to a few thousand high-quality data points for fine-tuning, while \cite{ji2023exploringimpactinstructiondata} suggests that using a significantly larger amount of data can yield better results. Given the capabilities of our base model, we determined that augmenting the fine-tuning phase with additional data was necessary. This strategy not only aids the model in mastering conversational techniques but also enriches it with supplementary knowledge. Consequently, We curated the following fine-tuning datasets: Infinity-Instruct Chinese data~\citep{InfinityInstruct2024} (approximately 700,000 entries, with all English data removed), multiple-choice questions from the Wanjuan-cn dataset~\citep{he2023wanjuan} which tested in both CoT and non-CoT formats, Ruozhiba data~\citep{better-ruozhiba}, self-awareness data with template modified from~\citep{2024EmoLLM}, and three English datasets: Code-Feedback~\citep{zheng2025opencodeinterpreterintegratingcodegeneration}, WebInstructSub~\citep{yue2024mammoth2} and OpenHermes-2.5~\citep{OpenHermes}. Further details are listed in \autoref{App:sft_data}.

We fine-tuned the pre-trained \model for approximately 4 epochs using the SFT dataset. The global batch size was set to 256, the maximum learning rate was set to $2 \times 10^{-5}$, and a cosine-annealing learning rate schedule was employed. The loss curve is shown in \autoref{App:sft_loss_curve}. Below, we present a series of ablation studies and evaluations to analyze the impact of different fine-tuning strategies on model performance.

\subsection{Ablation Studies and Evaluation}

To systematically evaluate the effectiveness of our fine-tuning approach, we conducted several experiments with varying data compositions and formats. The results are summarized in \autoref{tab:sft_results}, and the key findings are discussed below.

\begin{table}[ht]
    \centering
    \resizebox{\linewidth}{!}{
    \begin{tabular}{@{}lccc@{}}
        \toprule
        \textbf{Experiment} & \textbf{CEVAL Acc.} & \textbf{CMMLU Acc.} & \textbf{MMLU Acc.} \\
        \midrule
        Full Infinity-Instruct + Wanjuan MCQ & 32.35 & 26.32 & 25.50 \\
        700K Chinese Infinity-Instruct + Wanjuan MCQ & 38.57 & 33.48 & 23.26 \\
        Chinese + 100\% English Data & 39.21 & 33.20 & 26.73 \\
        Chinese + 20\% English Data (Balanced) & 40.43 & 35.86 & 26.75 \\
        Chinese + 20\% English + English MCQ & 41.90 & 36.08 & 30.82 \\
        \bottomrule
    \end{tabular}}
    \caption{Performance of different fine-tuning strategies.}
    \label{tab:sft_results}
\end{table}

\paragraph{Experiment 1: Full Infinity-Instruct + Wanjuan MCQ}
In this experiment, we fine-tuned the model using the entire Infinity-Instruct dataset (approximately 7,000,000 entries) and the full Wanjuan multiple-choice dataset. While the model's performance on CEVAL improved with increasing fine-tuning steps, the accuracy plateaued at around 32\% after 18,000 steps. This suggests that even though the Wanjuan data was included in the pretraining phase, the model still benefited from additional fine-tuning, likely due to its relatively small size.

\paragraph{Experiment 2: 700K Chinese Infinity-Instruct + Wanjuan MCQ}
To address the issue of language mismatch, we filtered out the English data from Infinity-Instruct and retained only the 700,000 Chinese entries. This adjustment significantly improved the model's performance, achieving 38\% accuracy on CEVAL and 33\% on CMMLU. This result highlights the importance of aligning the fine-tuning data distribution with the pretraining phase.

\paragraph{Experiment 3: Chinese + 100\% English Data}
To explore the impact of English data on the model's multilingual capabilities, we fine-tuned the model using a balanced mix of Chinese and English data (340,000 entries each). While the model's performance on Chinese benchmarks remained stable, its performance on MMLU improved slightly, indicating that English data can complement the model's existing knowledge without degrading its Chinese capabilities.

\paragraph{Experiment 4: Chinese + 20\% English Data (Balanced)}
In this experiment, we fine-tuned the model with a data distribution that closely matched the pretraining phase (80\% Chinese and 20\% English). This approach not only improved the model's performance on Chinese benchmarks (CEVAL: 39.21\% → 40.43\%; CMMLU: 33.2\% → 35.86\%) but also maintained its performance on MMLU. This suggests that maintaining a balanced data distribution during fine-tuning is crucial for preserving the model's multilingual capabilities.

\paragraph{Experiment 5: Chinese + 20\% English + English MCQ}
To further enhance the model's reasoning abilities, we introduced additional English multiple-choice questions from datasets such as OpenBookQA, AI2 ARC, and LogiQA. This experiment resulted in a noticeable improvement in MMLU accuracy (26.75\% → 30.82\%), while the performance on Chinese benchmarks remained stable. This indicates that incorporating domain-specific question-answering data can enhance the model's reasoning capabilities without compromising its existing knowledge.

\subsection{Learning from Human Preferences}
To align \model with human preferences, we employ the Direct Preference Optimization (DPO)\cite{rafailov2024directpreferenceoptimizationlanguage} algorithm and sorted response pairs for model optimization.

In the preference dataset, the ratio of Chinese to English stands at 4:1, consistent with that in the pre-training phase. The Chinese dataset is derived from ultrafeedback-chinese\footnote{\url{https://huggingface.co/datasets/opencsg/UltraFeedback-chinese/}}, and the English dataset is derived from ultrafeedback-binarized-preferences\footnote{\url{https://huggingface.co/datasets/argilla/ultrafeedback-binarized-preferences}}. Both the reference model and the objective model of the DPO algorithm are initialized with the SFT version of \model. We conducted training on the data for 3 epochs. The global batch size was set to 128, the maximum learning rate was set to $5 \times 10^{-6}$, the pref\_beta set to 0.1, and a cosine-annealing learning rate schedule was employed. The loss curve is shown in \autoref{App:dpo_loss_curve}.

\subsection{Discussion}

Our ablation studies reveal several key insights:
\begin{itemize}
    \item \textbf{Data Distribution Matters}: Fine-tuning with a data distribution that closely matches the pretraining phase leads to better performance on both Chinese and English benchmarks.
    \item \textbf{Small Models Benefit from Additional Data}: Even for small models, fine-tuning with a larger dataset can improve performance, particularly when the pretraining data is limited.
    \item \textbf{Balanced Multilingual Fine-Tuning}: Incorporating a small proportion of English data during fine-tuning can enhance the model's multilingual capabilities without degrading its performance on Chinese tasks.
    \item \textbf{Exam-Style Data Enhances Performance}: Including domain-specific question-answering data, such as multiple-choice questions, can improve the model's reasoning abilities and overall benchmark performance.
\end{itemize}

To contextualize the performance of \model, we compare it with several state-of-the-art models on the CEVAL and CMMLU benchmarks, as shown in \autoref{tab:ceval_cmmlu}. Our best-performing model, \model-Chat, achieves competitive results, outperforming models of similar scale such as Tiny-Llama-1.1B and Gemma-2b-it, and approaching the performance of larger models like CT-LLM-SFT-2B. While \model-Chat does not yet match the performance of significantly larger models like Qwen1.5-1.8B-Chat or Qwen-7B, its results demonstrate the effectiveness of our resource-efficient approach, particularly given the limited computational resources used for training.

\begin{table}[h]
\centering
\begin{tabular}{lcc}
\toprule
\textbf{Model} & \textbf{CEVAL} & \textbf{CMMLU} \\
\midrule
Tiny-Llama-1.1B \citep{zhang2024tinyllamaopensourcesmalllanguage} & 25.02 & 24.03 \\
MiniCPM-1.2B \citep{minicpm2024} & 49.14 & 46.81 \\
Qwen1.5-1.8B-Chat \citep{qwen} & 56.84 & 54.11 \\
Phi2(2B)\citep{phi2}  & 23.37 & 24.18 \\
Gemma-2b-it \citep{gemma_2024} & 32.30 & 33.07 \\
CT-LLM-SFT-2B~\citep{Du2024ChineseTL} & 41.54 & 41.48 \\
ChatGLM-6B \citep{glm2024chatglm} & 38.90 & 37.48 \\
Llama2-7B \cite{touvron2023llama2openfoundation} & 32.42 & 31.11 \\
OLMo-7B \citep{Groeneveld2023OLMo} & 35.18 & 35.55 \\
Gemma-7B \citep{gemma_2024}& 42.57 & 44.20 \\
MAP-Neo-7B \citep{zhang2024mapneo} & 56.97 & 55.01 \\
Llama2-13B \cite{touvron2023llama2openfoundation} & 37.32 & 37.06 \\
\midrule
\model-1B-Chat & 41.90 & 36.08 \\
\model-1B-Chat-DPO & 42.04 & 36.04 \\
\bottomrule
\end{tabular}
\caption{Performance comparison of models on CEVAL and CMMLU benchmarks.}
\label{tab:ceval_cmmlu}
\end{table}

In conclusion, our fine-tuning approach demonstrates that careful dataset selection and composition can significantly enhance the performance of small language models like \model. By balancing the inclusion of diverse data types and maintaining alignment with the pretraining distribution, we achieved competitive results on both Chinese and English benchmarks, positioning \model as a strong contender among resource-efficient LLMs.

\section{Conclusion}

This paper introduces \model, a fully open-source Chinese-centric language model developed with limited computational resources, achieving competitive performance on benchmarks such as CEVAL (41.90\%) and CMMLU (36.08\%). By leveraging innovative techniques like Soft Mixture of Experts and enhanced feed-forward networks, along with systematic training optimizations, we demonstrate that high-quality LLMs can be built efficiently. Our work provides complete transparency, releasing the training pipeline, datasets, and intermediate checkpoints, offering practical guidance for small-scale LLM development. All resources are made publicly available to foster collaboration and advance accessible language technologies.

\bibliography{iclr2025_conference}

\begin{thebibliography}{60}
\providecommand{\natexlab}[1]{#1}
\providecommand{\url}[1]{\texttt{#1}}
\expandafter\ifx\csname urlstyle\endcsname\relax
  \providecommand{\doi}[1]{doi: #1}\else
  \providecommand{\doi}{doi: \begingroup \urlstyle{rm}\Url}\fi

\bibitem[min(2024)]{minicpm2024}
Minicpm: Unveiling the potential of end-side large language models.
\newblock In \emph{OpenBMB Blog}, 2024.

\bibitem[Abdin et~al.(2023)Abdin, Aneja, Bubeck, Mendes, Chen, Giorno, Eldan, Gopi, Gunasekar, Javaheripi, Kauffmann, Lee, Li, Nguyen, de~Rosa, Saarikivi, Salim, Shah, Santacroce, Behl, Kalai, Wang, Ward, Witte, Zhang, and Zhang]{phi2}
Marah Abdin, Jyoti Aneja, Sebastien Bubeck, Caio César~Teodoro Mendes, Weizhu Chen, Allie~Del Giorno, Ronen Eldan, Sivakanth Gopi, Suriya Gunasekar, Mojan Javaheripi, Piero Kauffmann, Yin~Tat Lee, Yuanzhi Li, Anh Nguyen, Gustavo de~Rosa, Olli Saarikivi, Adil Salim, Shital Shah, Michael Santacroce, Harkirat~Singh Behl, Adam~Taumann Kalai, Xin Wang, Rachel Ward, Philipp Witte, Cyril Zhang, and Yi~Zhang.
\newblock Phi-2: The surprising power of small language models.
\newblock \url{https://www.microsoft.com/en-us/research/blog/phi-2-the-surprising-power-of-small-language-models/}, 2023.

\bibitem[AI et~al.(2024)AI, :, Young, Chen, Li, Huang, Zhang, Zhang, Li, Zhu, Chen, Chang, Yu, Liu, Liu, Yue, Yang, Yang, Yu, Xie, Huang, Hu, Ren, Niu, Nie, Xu, Liu, Wang, Cai, Gu, Liu, and Dai]{ai2024yiopenfoundationmodels}
01. AI, :, Alex Young, Bei Chen, Chao Li, Chengen Huang, Ge~Zhang, Guanwei Zhang, Heng Li, Jiangcheng Zhu, Jianqun Chen, Jing Chang, Kaidong Yu, Peng Liu, Qiang Liu, Shawn Yue, Senbin Yang, Shiming Yang, Tao Yu, Wen Xie, Wenhao Huang, Xiaohui Hu, Xiaoyi Ren, Xinyao Niu, Pengcheng Nie, Yuchi Xu, Yudong Liu, Yue Wang, Yuxuan Cai, Zhenyu Gu, Zhiyuan Liu, and Zonghong Dai.
\newblock Yi: Open foundation models by 01.ai, 2024.
\newblock URL \url{https://arxiv.org/abs/2403.04652}.

\bibitem[{BAAI}(2024)]{InfinityInstruct2024}
{BAAI}.
\newblock Infinity instruct.
\newblock \emph{arXiv preprint}, arXiv:2406.XXXX, 2024.

\bibitem[Bai et~al.(2023{\natexlab{a}})Bai, Bai, Chu, Cui, Dang, Deng, Fan, Ge, Han, Huang, Hui, Ji, Li, Lin, Lin, Liu, Liu, Lu, Lu, Ma, Men, Ren, Ren, Tan, Tan, Tu, Wang, Wang, Wang, Wu, Xu, Xu, Yang, Yang, Yang, Yang, Yao, Yu, Yuan, Yuan, Zhang, Zhang, Zhang, Zhang, Zhou, Zhou, Zhou, and Zhu]{bai2023qwentechnicalreport}
Jinze Bai, Shuai Bai, Yunfei Chu, Zeyu Cui, Kai Dang, Xiaodong Deng, Yang Fan, Wenbin Ge, Yu~Han, Fei Huang, Binyuan Hui, Luo Ji, Mei Li, Junyang Lin, Runji Lin, Dayiheng Liu, Gao Liu, Chengqiang Lu, Keming Lu, Jianxin Ma, Rui Men, Xingzhang Ren, Xuancheng Ren, Chuanqi Tan, Sinan Tan, Jianhong Tu, Peng Wang, Shijie Wang, Wei Wang, Shengguang Wu, Benfeng Xu, Jin Xu, An~Yang, Hao Yang, Jian Yang, Shusheng Yang, Yang Yao, Bowen Yu, Hongyi Yuan, Zheng Yuan, Jianwei Zhang, Xingxuan Zhang, Yichang Zhang, Zhenru Zhang, Chang Zhou, Jingren Zhou, Xiaohuan Zhou, and Tianhang Zhu.
\newblock Qwen technical report, 2023{\natexlab{a}}.
\newblock URL \url{https://arxiv.org/abs/2309.16609}.

\bibitem[Bai et~al.(2023{\natexlab{b}})Bai, Bai, Chu, Cui, Dang, Deng, Fan, Ge, Han, Huang, Hui, Ji, Li, Lin, Lin, Liu, Liu, Lu, Lu, Ma, Men, Ren, Ren, Tan, Tan, Tu, Wang, Wang, Wang, Wu, Xu, Xu, Yang, Yang, Yang, Yang, Yao, Yu, Yuan, Yuan, Zhang, Zhang, Zhang, Zhang, Zhou, Zhou, Zhou, and Zhu]{qwen}
Jinze Bai, Shuai Bai, Yunfei Chu, Zeyu Cui, Kai Dang, Xiaodong Deng, Yang Fan, Wenbin Ge, Yu~Han, Fei Huang, Binyuan Hui, Luo Ji, Mei Li, Junyang Lin, Runji Lin, Dayiheng Liu, Gao Liu, Chengqiang Lu, Keming Lu, Jianxin Ma, Rui Men, Xingzhang Ren, Xuancheng Ren, Chuanqi Tan, Sinan Tan, Jianhong Tu, Peng Wang, Shijie Wang, Wei Wang, Shengguang Wu, Benfeng Xu, Jin Xu, An~Yang, Hao Yang, Jian Yang, Shusheng Yang, Yang Yao, Bowen Yu, Hongyi Yuan, Zheng Yuan, Jianwei Zhang, Xingxuan Zhang, Yichang Zhang, Zhenru Zhang, Chang Zhou, Jingren Zhou, Xiaohuan Zhou, and Tianhang Zhu.
\newblock Qwen technical report.
\newblock \emph{arXiv preprint arXiv:2309.16609}, 2023{\natexlab{b}}.

\bibitem[BELLEGroup(2023)]{BELLE}
BELLEGroup.
\newblock Belle: Be everyone's large language model engine.
\newblock \url{https://github.com/LianjiaTech/BELLE}, 2023.

\bibitem[Cai et~al.(2024)Cai, Cao, Chen, Chen, Chen, Chen, Chen, Chen, Chen, Chu, Dong, Duan, Fan, Fei, Gao, Ge, Gu, Gu, Gui, Guo, Guo, He, Hu, Huang, Jiang, Jiao, Jin, Lei, Li, Li, Li, Li, Li, Li, Liu, Liu, Hong, Liu, Liu, Liu, Lv, Lv, Lv, Ma, Ma, Ma, Ning, Ouyang, Qiu, Qu, Shang, Shao, Song, Song, Sui, Sun, Sun, Tang, Wang, Wang, Wang, Wang, Wang, Wang, Wang, Wei, Weng, Wu, Xiong, Xu, Xu, Yan, Yan, Yang, Ye, Ying, Yu, Yu, Zang, Zhang, Zhang, Zhang, Zhang, Zhang, Zhang, Zhang, Zhang, Zhang, Zhang, Zhang, Zhao, Zhao, Zhao, Zhou, Zhou, Zhuo, Zou, Qiu, Qiao, and Lin]{cai2024internlm2}
Zheng Cai, Maosong Cao, Haojiong Chen, Kai Chen, Keyu Chen, Xin Chen, Xun Chen, Zehui Chen, Zhi Chen, Pei Chu, Xiaoyi Dong, Haodong Duan, Qi~Fan, Zhaoye Fei, Yang Gao, Jiaye Ge, Chenya Gu, Yuzhe Gu, Tao Gui, Aijia Guo, Qipeng Guo, Conghui He, Yingfan Hu, Ting Huang, Tao Jiang, Penglong Jiao, Zhenjiang Jin, Zhikai Lei, Jiaxing Li, Jingwen Li, Linyang Li, Shuaibin Li, Wei Li, Yining Li, Hongwei Liu, Jiangning Liu, Jiawei Hong, Kaiwen Liu, Kuikun Liu, Xiaoran Liu, Chengqi Lv, Haijun Lv, Kai Lv, Li~Ma, Runyuan Ma, Zerun Ma, Wenchang Ning, Linke Ouyang, Jiantao Qiu, Yuan Qu, Fukai Shang, Yunfan Shao, Demin Song, Zifan Song, Zhihao Sui, Peng Sun, Yu~Sun, Huanze Tang, Bin Wang, Guoteng Wang, Jiaqi Wang, Jiayu Wang, Rui Wang, Yudong Wang, Ziyi Wang, Xingjian Wei, Qizhen Weng, Fan Wu, Yingtong Xiong, Chao Xu, Ruiliang Xu, Hang Yan, Yirong Yan, Xiaogui Yang, Haochen Ye, Huaiyuan Ying, Jia Yu, Jing Yu, Yuhang Zang, Chuyu Zhang, Li~Zhang, Pan Zhang, Peng Zhang, Ruijie Zhang, Shuo Zhang, Songyang Zhang, Wenjian Zhang,
  Wenwei Zhang, Xingcheng Zhang, Xinyue Zhang, Hui Zhao, Qian Zhao, Xiaomeng Zhao, Fengzhe Zhou, Zaida Zhou, Jingming Zhuo, Yicheng Zou, Xipeng Qiu, Yu~Qiao, and Dahua Lin.
\newblock Internlm2 technical report, 2024.

\bibitem[Chowdhery et~al.(2022)Chowdhery, Narang, Devlin, Bosma, Mishra, Roberts, Barham, Chung, Sutton, Gehrmann, Schuh, Shi, Tsvyashchenko, Maynez, Rao, Barnes, Tay, Shazeer, Prabhakaran, Reif, Du, Hutchinson, Pope, Bradbury, Austin, Isard, Gur-Ari, Yin, Duke, Levskaya, Ghemawat, Dev, Michalewski, Garcia, Misra, Robinson, Fedus, Zhou, Ippolito, Luan, Lim, Zoph, Spiridonov, Sepassi, Dohan, Agrawal, Omernick, Dai, Pillai, Pellat, Lewkowycz, Moreira, Child, Polozov, Lee, Zhou, Wang, Saeta, Diaz, Firat, Catasta, Wei, Meier-Hellstern, Eck, Dean, Petrov, and Fiedel]{chowdhery2022palmscalinglanguagemodeling}
Aakanksha Chowdhery, Sharan Narang, Jacob Devlin, Maarten Bosma, Gaurav Mishra, Adam Roberts, Paul Barham, Hyung~Won Chung, Charles Sutton, Sebastian Gehrmann, Parker Schuh, Kensen Shi, Sasha Tsvyashchenko, Joshua Maynez, Abhishek Rao, Parker Barnes, Yi~Tay, Noam Shazeer, Vinodkumar Prabhakaran, Emily Reif, Nan Du, Ben Hutchinson, Reiner Pope, James Bradbury, Jacob Austin, Michael Isard, Guy Gur-Ari, Pengcheng Yin, Toju Duke, Anselm Levskaya, Sanjay Ghemawat, Sunipa Dev, Henryk Michalewski, Xavier Garcia, Vedant Misra, Kevin Robinson, Liam Fedus, Denny Zhou, Daphne Ippolito, David Luan, Hyeontaek Lim, Barret Zoph, Alexander Spiridonov, Ryan Sepassi, David Dohan, Shivani Agrawal, Mark Omernick, Andrew~M. Dai, Thanumalayan~Sankaranarayana Pillai, Marie Pellat, Aitor Lewkowycz, Erica Moreira, Rewon Child, Oleksandr Polozov, Katherine Lee, Zongwei Zhou, Xuezhi Wang, Brennan Saeta, Mark Diaz, Orhan Firat, Michele Catasta, Jason Wei, Kathy Meier-Hellstern, Douglas Eck, Jeff Dean, Slav Petrov, and Noah Fiedel.
\newblock Palm: Scaling language modeling with pathways, 2022.
\newblock URL \url{https://arxiv.org/abs/2204.02311}.

\bibitem[Dai et~al.(2024)Dai, Deng, Zhao, Xu, Gao, Chen, Li, Zeng, Yu, Wu, Xie, Li, Huang, Luo, Ruan, Sui, and Liang]{dai2024deepseekmoeultimateexpertspecialization}
Damai Dai, Chengqi Deng, Chenggang Zhao, R.~X. Xu, Huazuo Gao, Deli Chen, Jiashi Li, Wangding Zeng, Xingkai Yu, Y.~Wu, Zhenda Xie, Y.~K. Li, Panpan Huang, Fuli Luo, Chong Ruan, Zhifang Sui, and Wenfeng Liang.
\newblock Deepseekmoe: Towards ultimate expert specialization in mixture-of-experts language models, 2024.
\newblock URL \url{https://arxiv.org/abs/2401.06066}.

\bibitem[Dao et~al.(2022)Dao, Fu, Ermon, Rudra, and Ré]{dao2022flashattentionfastmemoryefficientexact}
Tri Dao, Daniel~Y. Fu, Stefano Ermon, Atri Rudra, and Christopher Ré.
\newblock Flashattention: Fast and memory-efficient exact attention with io-awareness, 2022.
\newblock URL \url{https://arxiv.org/abs/2205.14135}.

\bibitem[DeepSeek-AI et~al.(2024{\natexlab{a}})DeepSeek-AI, :, Bi, Chen, Chen, Chen, Dai, Deng, Ding, Dong, Du, Fu, Gao, Gao, Gao, Ge, Guan, Guo, Guo, Hao, Hao, He, Hu, Huang, Li, Li, Li, Li, Li, Liang, Lin, Liu, Liu, Liu, Liu, Liu, Liu, Lu, Lu, Luo, Ma, Nie, Pei, Piao, Qiu, Qu, Ren, Ren, Ruan, Sha, Shao, Song, Su, Sun, Sun, Tang, Wang, Wang, Wang, Wang, Wang, Wu, Wu, Xie, Xie, Xie, Xiong, Xu, Xu, Xu, Yang, You, Yu, Yu, Zhang, Zhang, Zhang, Zhang, Zhang, Zhang, Zhang, Zhang, Zhao, Zhao, Zhou, Zhou, Zhu, and Zou]{deepseekai2024deepseek}
DeepSeek-AI, :, Xiao Bi, Deli Chen, Guanting Chen, Shanhuang Chen, Damai Dai, Chengqi Deng, Honghui Ding, Kai Dong, Qiushi Du, Zhe Fu, Huazuo Gao, Kaige Gao, Wenjun Gao, Ruiqi Ge, Kang Guan, Daya Guo, Jianzhong Guo, Guangbo Hao, Zhewen Hao, Ying He, Wenjie Hu, Panpan Huang, Erhang Li, Guowei Li, Jiashi Li, Yao Li, Y.~K. Li, Wenfeng Liang, Fangyun Lin, A.~X. Liu, Bo~Liu, Wen Liu, Xiaodong Liu, Xin Liu, Yiyuan Liu, Haoyu Lu, Shanghao Lu, Fuli Luo, Shirong Ma, Xiaotao Nie, Tian Pei, Yishi Piao, Junjie Qiu, Hui Qu, Tongzheng Ren, Zehui Ren, Chong Ruan, Zhangli Sha, Zhihong Shao, Junxiao Song, Xuecheng Su, Jingxiang Sun, Yaofeng Sun, Minghui Tang, Bingxuan Wang, Peiyi Wang, Shiyu Wang, Yaohui Wang, Yongji Wang, Tong Wu, Y.~Wu, Xin Xie, Zhenda Xie, Ziwei Xie, Yiliang Xiong, Hanwei Xu, R.~X. Xu, Yanhong Xu, Dejian Yang, Yuxiang You, Shuiping Yu, Xingkai Yu, B.~Zhang, Haowei Zhang, Lecong Zhang, Liyue Zhang, Mingchuan Zhang, Minghua Zhang, Wentao Zhang, Yichao Zhang, Chenggang Zhao, Yao Zhao, Shangyan Zhou, Shunfeng
  Zhou, Qihao Zhu, and Yuheng Zou.
\newblock Deepseek llm: Scaling open-source language models with longtermism, 2024{\natexlab{a}}.

\bibitem[DeepSeek-AI et~al.(2024{\natexlab{b}})DeepSeek-AI, Liu, Feng, Wang, Wang, Liu, Zhao, Dengr, Ruan, Dai, Guo, Yang, Chen, Ji, Li, Lin, Luo, Hao, Chen, Li, Zhang, Xu, Yang, Zhang, Ding, Xin, Gao, Li, Qu, Cai, Liang, Guo, Ni, Li, Chen, Yuan, Qiu, Song, Dong, Gao, Guan, Wang, Zhang, Xu, Xia, Zhao, Zhang, Li, Wang, Zhang, Zhang, Tang, Li, Tian, Huang, Wang, Zhang, Zhu, Chen, Du, Chen, Jin, Ge, Pan, Xu, Chen, Li, Lu, Zhou, Chen, Wu, Ye, Ma, Wang, Zhou, Yu, Zhou, Zheng, Wang, Pei, Yuan, Sun, Xiao, Zeng, An, Liu, Liang, Gao, Zhang, Li, Jin, Wang, Bi, Liu, Wang, Shen, Chen, Chen, Nie, Sun, Wang, Liu, Xie, Yu, Song, Zhou, Yang, Lu, Su, Wu, Li, Wei, Zhu, Xu, Huang, Li, Zhao, Sun, Li, Wang, Zheng, Zhang, Xiong, Zhao, He, Tang, Piao, Dong, Tan, Liu, Wang, Guo, Zhu, Wang, Zou, Zha, Ma, Yan, You, Liu, Ren, Ren, Sha, Fu, Huang, Zhang, Xie, Hao, Shao, Wen, Xu, Zhang, Li, Wang, Gu, Li, and Xie]{deepseekai2024deepseekv2strongeconomicalefficient}
DeepSeek-AI, Aixin Liu, Bei Feng, Bin Wang, Bingxuan Wang, Bo~Liu, Chenggang Zhao, Chengqi Dengr, Chong Ruan, Damai Dai, Daya Guo, Dejian Yang, Deli Chen, Dongjie Ji, Erhang Li, Fangyun Lin, Fuli Luo, Guangbo Hao, Guanting Chen, Guowei Li, H.~Zhang, Hanwei Xu, Hao Yang, Haowei Zhang, Honghui Ding, Huajian Xin, Huazuo Gao, Hui Li, Hui Qu, J.~L. Cai, Jian Liang, Jianzhong Guo, Jiaqi Ni, Jiashi Li, Jin Chen, Jingyang Yuan, Junjie Qiu, Junxiao Song, Kai Dong, Kaige Gao, Kang Guan, Lean Wang, Lecong Zhang, Lei Xu, Leyi Xia, Liang Zhao, Liyue Zhang, Meng Li, Miaojun Wang, Mingchuan Zhang, Minghua Zhang, Minghui Tang, Mingming Li, Ning Tian, Panpan Huang, Peiyi Wang, Peng Zhang, Qihao Zhu, Qinyu Chen, Qiushi Du, R.~J. Chen, R.~L. Jin, Ruiqi Ge, Ruizhe Pan, Runxin Xu, Ruyi Chen, S.~S. Li, Shanghao Lu, Shangyan Zhou, Shanhuang Chen, Shaoqing Wu, Shengfeng Ye, Shirong Ma, Shiyu Wang, Shuang Zhou, Shuiping Yu, Shunfeng Zhou, Size Zheng, T.~Wang, Tian Pei, Tian Yuan, Tianyu Sun, W.~L. Xiao, Wangding Zeng, Wei An, Wen
  Liu, Wenfeng Liang, Wenjun Gao, Wentao Zhang, X.~Q. Li, Xiangyue Jin, Xianzu Wang, Xiao Bi, Xiaodong Liu, Xiaohan Wang, Xiaojin Shen, Xiaokang Chen, Xiaosha Chen, Xiaotao Nie, Xiaowen Sun, Xiaoxiang Wang, Xin Liu, Xin Xie, Xingkai Yu, Xinnan Song, Xinyi Zhou, Xinyu Yang, Xuan Lu, Xuecheng Su, Y.~Wu, Y.~K. Li, Y.~X. Wei, Y.~X. Zhu, Yanhong Xu, Yanping Huang, Yao Li, Yao Zhao, Yaofeng Sun, Yaohui Li, Yaohui Wang, Yi~Zheng, Yichao Zhang, Yiliang Xiong, Yilong Zhao, Ying He, Ying Tang, Yishi Piao, Yixin Dong, Yixuan Tan, Yiyuan Liu, Yongji Wang, Yongqiang Guo, Yuchen Zhu, Yuduan Wang, Yuheng Zou, Yukun Zha, Yunxian Ma, Yuting Yan, Yuxiang You, Yuxuan Liu, Z.~Z. Ren, Zehui Ren, Zhangli Sha, Zhe Fu, Zhen Huang, Zhen Zhang, Zhenda Xie, Zhewen Hao, Zhihong Shao, Zhiniu Wen, Zhipeng Xu, Zhongyu Zhang, Zhuoshu Li, Zihan Wang, Zihui Gu, Zilin Li, and Ziwei Xie.
\newblock Deepseek-v2: A strong, economical, and efficient mixture-of-experts language model, 2024{\natexlab{b}}.
\newblock URL \url{https://arxiv.org/abs/2405.04434}.

\bibitem[Du et~al.(2024)Du, Yu, Gao, Pan, Cheng, Ma, Yuan, Qu, Liu, Zheng, Luo, Zhou, Yuan, Chen, Fu, and Zhang]{Du2024ChineseTL}
Xinrun Du, Zhouliang Yu, Songyang Gao, Ding Pan, Yuyang Cheng, Ziyang Ma, Ruibin Yuan, Xingwei Qu, Jiaheng Liu, Tianyu Zheng, Xinchen Luo, Guorui Zhou, Binhang Yuan, Wenhu Chen, Jie Fu, and Ge~Zhang.
\newblock Chinese tiny llm: Pretraining a chinese-centric large language model.
\newblock \emph{ArXiv}, abs/2404.04167, 2024.
\newblock URL \url{https://api.semanticscholar.org/CorpusID:268987532}.

\bibitem[Fedus et~al.(2022)Fedus, Zoph, and Shazeer]{fedus2022switchtransformersscalingtrillion}
William Fedus, Barret Zoph, and Noam Shazeer.
\newblock Switch transformers: Scaling to trillion parameter models with simple and efficient sparsity, 2022.
\newblock URL \url{https://arxiv.org/abs/2101.03961}.

\bibitem[Gemma~Team et~al.(2024)Gemma~Team, Hardin, Dadashi, Bhupatiraju, Sifre, Rivière, Kale, Love, Tafti, Hussenot, and et~al.]{gemma_2024}
Thomas~Mesnard Gemma~Team, Cassidy Hardin, Robert Dadashi, Surya Bhupatiraju, Laurent Sifre, Morgane Rivière, Mihir~Sanjay Kale, Juliette Love, Pouya Tafti, Léonard Hussenot, and et~al.
\newblock Gemma.
\newblock 2024.
\newblock \doi{10.34740/KAGGLE/M/3301}.
\newblock URL \url{https://www.kaggle.com/m/3301}.

\bibitem[GLM et~al.(2024)GLM, Zeng, Xu, Wang, Zhang, Yin, Rojas, Feng, Zhao, Lai, Yu, Wang, Sun, Zhang, Cheng, Gui, Tang, Zhang, Li, Zhao, Wu, Zhong, Liu, Huang, Zhang, Zheng, Lu, Duan, Zhang, Cao, Yang, Tam, Zhao, Liu, Xia, Zhang, Gu, Lv, Liu, Liu, Yang, Song, Zhang, An, Xu, Niu, Yang, Li, Bai, Dong, Qi, Wang, Yang, Du, Hou, and Wang]{glm2024chatglm}
Team GLM, Aohan Zeng, Bin Xu, Bowen Wang, Chenhui Zhang, Da~Yin, Diego Rojas, Guanyu Feng, Hanlin Zhao, Hanyu Lai, Hao Yu, Hongning Wang, Jiadai Sun, Jiajie Zhang, Jiale Cheng, Jiayi Gui, Jie Tang, Jing Zhang, Juanzi Li, Lei Zhao, Lindong Wu, Lucen Zhong, Mingdao Liu, Minlie Huang, Peng Zhang, Qinkai Zheng, Rui Lu, Shuaiqi Duan, Shudan Zhang, Shulin Cao, Shuxun Yang, Weng~Lam Tam, Wenyi Zhao, Xiao Liu, Xiao Xia, Xiaohan Zhang, Xiaotao Gu, Xin Lv, Xinghan Liu, Xinyi Liu, Xinyue Yang, Xixuan Song, Xunkai Zhang, Yifan An, Yifan Xu, Yilin Niu, Yuantao Yang, Yueyan Li, Yushi Bai, Yuxiao Dong, Zehan Qi, Zhaoyu Wang, Zhen Yang, Zhengxiao Du, Zhenyu Hou, and Zihan Wang.
\newblock Chatglm: A family of large language models from glm-130b to glm-4 all tools, 2024.

\bibitem[Groeneveld et~al.(2024{\natexlab{a}})Groeneveld, Beltagy, Walsh, Bhagia, Kinney, Tafjord, Jha, Ivison, Magnusson, Wang, Arora, Atkinson, Authur, Chandu, Cohan, Dumas, Elazar, Gu, Hessel, Khot, Merrill, Morrison, Muennighoff, Naik, Nam, Peters, Pyatkin, Ravichander, Schwenk, Shah, Smith, Strubell, Subramani, Wortsman, Dasigi, Lambert, Richardson, Zettlemoyer, Dodge, Lo, Soldaini, Smith, and Hajishirzi]{OLMo}
Dirk Groeneveld, Iz~Beltagy, Pete Walsh, Akshita Bhagia, Rodney Kinney, Oyvind Tafjord, A.~Jha, Hamish Ivison, Ian Magnusson, Yizhong Wang, Shane Arora, David Atkinson, Russell Authur, Khyathi~Raghavi Chandu, Arman Cohan, Jennifer Dumas, Yanai Elazar, Yuling Gu, Jack Hessel, Tushar Khot, William Merrill, Jacob~Daniel Morrison, Niklas Muennighoff, Aakanksha Naik, Crystal Nam, Matthew~E. Peters, Valentina Pyatkin, Abhilasha Ravichander, Dustin Schwenk, Saurabh Shah, Will Smith, Emma Strubell, Nishant Subramani, Mitchell Wortsman, Pradeep Dasigi, Nathan Lambert, Kyle Richardson, Luke Zettlemoyer, Jesse Dodge, Kyle Lo, Luca Soldaini, Noah~A. Smith, and Hanna Hajishirzi.
\newblock Olmo: Accelerating the science of language models.
\newblock \emph{arXiv preprint}, 2024{\natexlab{a}}.
\newblock URL \url{https://api.semanticscholar.org/CorpusID:267365485}.

\bibitem[Groeneveld et~al.(2024{\natexlab{b}})Groeneveld, Beltagy, Walsh, Bhagia, Kinney, Tafjord, Jha, Ivison, Magnusson, Wang, Arora, Atkinson, Authur, Chandu, Cohan, Dumas, Elazar, Gu, Hessel, Khot, Merrill, Morrison, Muennighoff, Naik, Nam, Peters, Pyatkin, Ravichander, Schwenk, Shah, Smith, Subramani, Wortsman, Dasigi, Lambert, Richardson, Dodge, Lo, Soldaini, Smith, and Hajishirzi]{Groeneveld2023OLMo}
Dirk Groeneveld, Iz~Beltagy, Pete Walsh, Akshita Bhagia, Rodney Kinney, Oyvind Tafjord, Ananya~Harsh Jha, Hamish Ivison, Ian Magnusson, Yizhong Wang, Shane Arora, David Atkinson, Russell Authur, Khyathi Chandu, Arman Cohan, Jennifer Dumas, Yanai Elazar, Yuling Gu, Jack Hessel, Tushar Khot, William Merrill, Jacob Morrison, Niklas Muennighoff, Aakanksha Naik, Crystal Nam, Matthew~E. Peters, Valentina Pyatkin, Abhilasha Ravichander, Dustin Schwenk, Saurabh Shah, Will Smith, Nishant Subramani, Mitchell Wortsman, Pradeep Dasigi, Nathan Lambert, Kyle Richardson, Jesse Dodge, Kyle Lo, Luca Soldaini, Noah~A. Smith, and Hannaneh Hajishirzi.
\newblock Olmo: Accelerating the science of language models.
\newblock \emph{Preprint}, 2024{\natexlab{b}}.

\bibitem[He et~al.(2023)He, Jin, Xu, Qiu, Wang, Li, Yan, Wang, and Lin]{he2023wanjuan}
Conghui He, Zhenjiang Jin, Chao Xu, Jiantao Qiu, Bin Wang, Wei Li, Hang Yan, Jiaqi Wang, and Dahua Lin.
\newblock Wanjuan: A comprehensive multimodal dataset for advancing english and chinese large models, 2023.

\bibitem[Jacobs et~al.(1991)Jacobs, Jordan, Nowlan, and Hinton]{6797059}
Robert~A. Jacobs, Michael~I. Jordan, Steven~J. Nowlan, and Geoffrey~E. Hinton.
\newblock Adaptive mixtures of local experts.
\newblock \emph{Neural Computation}, 3\penalty0 (1):\penalty0 79--87, 1991.
\newblock \doi{10.1162/neco.1991.3.1.79}.

\bibitem[Ji et~al.(2023)Ji, Deng, Gong, Peng, Niu, Zhang, Ma, and Li]{ji2023exploringimpactinstructiondata}
Yunjie Ji, Yong Deng, Yan Gong, Yiping Peng, Qiang Niu, Lei Zhang, Baochang Ma, and Xiangang Li.
\newblock Exploring the impact of instruction data scaling on large language models: An empirical study on real-world use cases, 2023.
\newblock URL \url{https://arxiv.org/abs/2303.14742}.

\bibitem[Jiang et~al.(2023)Jiang, Sablayrolles, Mensch, Bamford, Chaplot, de~las Casas, Bressand, Lengyel, Lample, Saulnier, Lavaud, Lachaux, Stock, Scao, Lavril, Wang, Lacroix, and Sayed]{jiang2023mistral}
Albert~Q. Jiang, Alexandre Sablayrolles, Arthur Mensch, Chris Bamford, Devendra~Singh Chaplot, Diego de~las Casas, Florian Bressand, Gianna Lengyel, Guillaume Lample, Lucile Saulnier, Lélio~Renard Lavaud, Marie-Anne Lachaux, Pierre Stock, Teven~Le Scao, Thibaut Lavril, Thomas Wang, Timothée Lacroix, and William~El Sayed.
\newblock Mistral 7b, 2023.

\bibitem[Kalamkar et~al.(2019)Kalamkar, Mudigere, Mellempudi, Das, Banerjee, Avancha, Vooturi, Jammalamadaka, Huang, Yuen, Yang, Park, Heinecke, Georganas, Srinivasan, Kundu, Smelyanskiy, Kaul, and Dubey]{kalamkar2019studybfloat16deeplearning}
Dhiraj Kalamkar, Dheevatsa Mudigere, Naveen Mellempudi, Dipankar Das, Kunal Banerjee, Sasikanth Avancha, Dharma~Teja Vooturi, Nataraj Jammalamadaka, Jianyu Huang, Hector Yuen, Jiyan Yang, Jongsoo Park, Alexander Heinecke, Evangelos Georganas, Sudarshan Srinivasan, Abhisek Kundu, Misha Smelyanskiy, Bharat Kaul, and Pradeep Dubey.
\newblock A study of bfloat16 for deep learning training, 2019.
\newblock URL \url{https://arxiv.org/abs/1905.12322}.

\bibitem[Kingma \& Ba(2017)Kingma and Ba]{kingma2017adammethodstochasticoptimization}
Diederik~P. Kingma and Jimmy Ba.
\newblock Adam: A method for stochastic optimization, 2017.
\newblock URL \url{https://arxiv.org/abs/1412.6980}.

\bibitem[Lepikhin et~al.(2021)Lepikhin, Lee, Xu, Chen, Firat, Huang, Krikun, Shazeer, and Chen]{lepikhin2021gshard}
Dmitry Lepikhin, HyoukJoong Lee, Yuanzhong Xu, Dehao Chen, Orhan Firat, Yanping Huang, Maxim Krikun, Noam Shazeer, and Zhifeng Chen.
\newblock {\{}GS{\}}hard: Scaling giant models with conditional computation and automatic sharding.
\newblock In \emph{International Conference on Learning Representations}, 2021.
\newblock URL \url{https://openreview.net/forum?id=qrwe7XHTmYb}.

\bibitem[Li et~al.(2023)Li, Allal, Zi, Muennighoff, Kocetkov, Mou, Marone, Akiki, Li, Chim, Liu, Zheltonozhskii, Zhuo, Wang, Dehaene, Davaadorj, Lamy-Poirier, Monteiro, Shliazhko, Gontier, Meade, Zebaze, Yee, Umapathi, Zhu, Lipkin, Oblokulov, Wang, Murthy, Stillerman, Patel, Abulkhanov, Zocca, Dey, Zhang, Fahmy, Bhattacharyya, Yu, Singh, Luccioni, Villegas, Kunakov, Zhdanov, Romero, Lee, Timor, Ding, Schlesinger, Schoelkopf, Ebert, Dao, Mishra, Gu, Robinson, Anderson, Dolan-Gavitt, Contractor, Reddy, Fried, Bahdanau, Jernite, Ferrandis, Hughes, Wolf, Guha, von Werra, and de~Vries]{li2023starcoder}
Raymond Li, Loubna~Ben Allal, Yangtian Zi, Niklas Muennighoff, Denis Kocetkov, Chenghao Mou, Marc Marone, Christopher Akiki, Jia Li, Jenny Chim, Qian Liu, Evgenii Zheltonozhskii, Terry~Yue Zhuo, Thomas Wang, Olivier Dehaene, Mishig Davaadorj, Joel Lamy-Poirier, João Monteiro, Oleh Shliazhko, Nicolas Gontier, Nicholas Meade, Armel Zebaze, Ming-Ho Yee, Logesh~Kumar Umapathi, Jian Zhu, Benjamin Lipkin, Muhtasham Oblokulov, Zhiruo Wang, Rudra Murthy, Jason Stillerman, Siva~Sankalp Patel, Dmitry Abulkhanov, Marco Zocca, Manan Dey, Zhihan Zhang, Nour Fahmy, Urvashi Bhattacharyya, Wenhao Yu, Swayam Singh, Sasha Luccioni, Paulo Villegas, Maxim Kunakov, Fedor Zhdanov, Manuel Romero, Tony Lee, Nadav Timor, Jennifer Ding, Claire Schlesinger, Hailey Schoelkopf, Jan Ebert, Tri Dao, Mayank Mishra, Alex Gu, Jennifer Robinson, Carolyn~Jane Anderson, Brendan Dolan-Gavitt, Danish Contractor, Siva Reddy, Daniel Fried, Dzmitry Bahdanau, Yacine Jernite, Carlos~Muñoz Ferrandis, Sean Hughes, Thomas Wolf, Arjun Guha, Leandro von
  Werra, and Harm de~Vries.
\newblock Starcoder: may the source be with you!
\newblock 2023.

\bibitem[Liu et~al.(2023)Liu, Qiao, Neiswanger, Wang, Tan, Tao, Li, Wang, Sun, Pangarkar, et~al.]{liu2023llm360}
Zhengzhong Liu, Aurick Qiao, Willie Neiswanger, Hongyi Wang, Bowen Tan, Tianhua Tao, Junbo Li, Yuqi Wang, Suqi Sun, Omkar Pangarkar, et~al.
\newblock Llm360: Towards fully transparent open-source llms.
\newblock \emph{arXiv preprint arXiv:2312.06550}, 2023.

\bibitem[Ma et~al.(2018)Ma, Zhao, Yi, Chen, Hong, and Chi]{ma2018modeling}
Jiaqi Ma, Zhe Zhao, Xinyang Yi, Jilin Chen, Lichan Hong, and Ed~H Chi.
\newblock Modeling task relationships in multi-task learning with multi-gate mixture-of-experts.
\newblock In \emph{Proceedings of the 24th ACM SIGKDD international conference on knowledge discovery \& data mining}, pp.\  1930--1939, 2018.

\bibitem[Muennighoff et~al.(2024)Muennighoff, Soldaini, Groeneveld, Lo, Morrison, Min, Shi, Walsh, Tafjord, Lambert, et~al.]{muennighoff2024olmoe}
Niklas Muennighoff, Luca Soldaini, Dirk Groeneveld, Kyle Lo, Jacob Morrison, Sewon Min, Weijia Shi, Pete Walsh, Oyvind Tafjord, Nathan Lambert, et~al.
\newblock Olmoe: Open mixture-of-experts language models.
\newblock \emph{arXiv preprint arXiv:2409.02060}, 2024.

\bibitem[OLMo et~al.(2024)OLMo, Walsh, Soldaini, Groeneveld, Lo, Arora, Bhagia, Gu, Huang, Jordan, et~al.]{olmo20242}
Team OLMo, Pete Walsh, Luca Soldaini, Dirk Groeneveld, Kyle Lo, Shane Arora, Akshita Bhagia, Yuling Gu, Shengyi Huang, Matt Jordan, et~al.
\newblock 2 olmo 2 furious.
\newblock \emph{arXiv preprint arXiv:2501.00656}, 2024.

\bibitem[Puigcerver et~al.(2024)Puigcerver, Riquelme, Mustafa, and Houlsby]{puigcerver2024sparsesoftmixturesexperts}
Joan Puigcerver, Carlos Riquelme, Basil Mustafa, and Neil Houlsby.
\newblock From sparse to soft mixtures of experts, 2024.
\newblock URL \url{https://arxiv.org/abs/2308.00951}.

\bibitem[Rafailov et~al.(2024)Rafailov, Sharma, Mitchell, Ermon, Manning, and Finn]{rafailov2024directpreferenceoptimizationlanguage}
Rafael Rafailov, Archit Sharma, Eric Mitchell, Stefano Ermon, Christopher~D. Manning, and Chelsea Finn.
\newblock Direct preference optimization: Your language model is secretly a reward model, 2024.
\newblock URL \url{https://arxiv.org/abs/2305.18290}.

\bibitem[Rivest(1992)]{RFC1321}
R.~Rivest.
\newblock The md5 message-digest algorithm.
\newblock RFC 1321, 4 1992.

\bibitem[Ruozhiba(2024)]{better-ruozhiba}
Misdirection Ruozhiba, FunnySaltyFish.
\newblock Better ruozhiba.
\newblock \url{https://github.com/FunnySaltyFish/Better-Ruozhiba}, 2024.

\bibitem[Shazeer(2020)]{shazeer2020gluvariantsimprovetransformer}
Noam Shazeer.
\newblock Glu variants improve transformer, 2020.
\newblock URL \url{https://arxiv.org/abs/2002.05202}.

\bibitem[Su et~al.(2023)Su, Lu, Pan, Murtadha, Wen, and Liu]{su2023roformerenhancedtransformerrotary}
Jianlin Su, Yu~Lu, Shengfeng Pan, Ahmed Murtadha, Bo~Wen, and Yunfeng Liu.
\newblock Roformer: Enhanced transformer with rotary position embedding, 2023.
\newblock URL \url{https://arxiv.org/abs/2104.09864}.

\bibitem[Sun et~al.(2024)Sun, Zhang, He, Li, Cheng, Liu, Yan, Shao, Tang, Zhang, Zhao, Chen, Zheng, Zhou, Li, Zhan, Zhou, Li, Yang, Wu, Yin, Huang, Jiang, and Qiu]{Sun2024MOSS}
Tianxiang Sun, Xiaotian Zhang, Zhengfu He, Peng Li, Qinyuan Cheng, Xiangyang Liu, Hang Yan, Yunfan Shao, Qiong Tang, Shiduo Zhang, Xingjian Zhao, Ke~Chen, Yining Zheng, Zhejian Zhou, Ruixiao Li, Jun Zhan, Yunhua Zhou, Linyang Li, Xiaogui Yang, Lingling Wu, Zhangyue Yin, Xuanjing Huang, Yu-Gang Jiang, and Xipeng Qiu.
\newblock Moss: An open conversational large language model.
\newblock \emph{Machine Intelligence Research}, 2024.
\newblock ISSN 2731-5398.
\newblock URL \url{https://github.com/OpenMOSS/MOSS}.

\bibitem[Tan et~al.(2024)Tan, Wang37, Neiswanger, Tao, Li, Koto, Wang, Sun, Pangarkar, Fan, et~al.]{tan2024llm360}
Bowen Tan, Hongyi Wang37, Willie Neiswanger, Tianhua Tao, Haonan Li, Fajri Koto, Yuqi Wang, Suqi Sun, Omkar Pangarkar, Richard Fan, et~al.
\newblock Llm360 k2-65b: Scaling up fully transparent open-source llms.
\newblock 2024.

\bibitem[Team(2024{\natexlab{a}})]{2024EmoLLM}
EmoLLM Team.
\newblock Emollm: Reinventing mental health support with large language models.
\newblock \url{https://github.com/SmartFlowAI/EmoLLM}, 2024{\natexlab{a}}.

\bibitem[Team et~al.(2024)Team, Mesnard, Hardin, Dadashi, Bhupatiraju, Pathak, Sifre, Rivière, Kale, Love, Tafti, Hussenot, Sessa, Chowdhery, Roberts, Barua, Botev, Castro-Ros, Slone, Héliou, Tacchetti, Bulanova, Paterson, Tsai, Shahriari, Lan, Choquette-Choo, Crepy, Cer, Ippolito, Reid, Buchatskaya, Ni, Noland, Yan, Tucker, Muraru, Rozhdestvenskiy, Michalewski, Tenney, Grishchenko, Austin, Keeling, Labanowski, Lespiau, Stanway, Brennan, Chen, Ferret, Chiu, Mao-Jones, Lee, Yu, Millican, Sjoesund, Lee, Dixon, Reid, Mikuła, Wirth, Sharman, Chinaev, Thain, Bachem, Chang, Wahltinez, Bailey, Michel, Yotov, Chaabouni, Comanescu, Jana, Anil, McIlroy, Liu, Mullins, Smith, Borgeaud, Girgin, Douglas, Pandya, Shakeri, De, Klimenko, Hennigan, Feinberg, Stokowiec, hui Chen, Ahmed, Gong, Warkentin, Peran, Giang, Farabet, Vinyals, Dean, Kavukcuoglu, Hassabis, Ghahramani, Eck, Barral, Pereira, Collins, Joulin, Fiedel, Senter, Andreev, and Kenealy]{gemmateam2024gemmaopenmodelsbased}
Gemma Team, Thomas Mesnard, Cassidy Hardin, Robert Dadashi, Surya Bhupatiraju, Shreya Pathak, Laurent Sifre, Morgane Rivière, Mihir~Sanjay Kale, Juliette Love, Pouya Tafti, Léonard Hussenot, Pier~Giuseppe Sessa, Aakanksha Chowdhery, Adam Roberts, Aditya Barua, Alex Botev, Alex Castro-Ros, Ambrose Slone, Amélie Héliou, Andrea Tacchetti, Anna Bulanova, Antonia Paterson, Beth Tsai, Bobak Shahriari, Charline~Le Lan, Christopher~A. Choquette-Choo, Clément Crepy, Daniel Cer, Daphne Ippolito, David Reid, Elena Buchatskaya, Eric Ni, Eric Noland, Geng Yan, George Tucker, George-Christian Muraru, Grigory Rozhdestvenskiy, Henryk Michalewski, Ian Tenney, Ivan Grishchenko, Jacob Austin, James Keeling, Jane Labanowski, Jean-Baptiste Lespiau, Jeff Stanway, Jenny Brennan, Jeremy Chen, Johan Ferret, Justin Chiu, Justin Mao-Jones, Katherine Lee, Kathy Yu, Katie Millican, Lars~Lowe Sjoesund, Lisa Lee, Lucas Dixon, Machel Reid, Maciej Mikuła, Mateo Wirth, Michael Sharman, Nikolai Chinaev, Nithum Thain, Olivier Bachem,
  Oscar Chang, Oscar Wahltinez, Paige Bailey, Paul Michel, Petko Yotov, Rahma Chaabouni, Ramona Comanescu, Reena Jana, Rohan Anil, Ross McIlroy, Ruibo Liu, Ryan Mullins, Samuel~L Smith, Sebastian Borgeaud, Sertan Girgin, Sholto Douglas, Shree Pandya, Siamak Shakeri, Soham De, Ted Klimenko, Tom Hennigan, Vlad Feinberg, Wojciech Stokowiec, Yu~hui Chen, Zafarali Ahmed, Zhitao Gong, Tris Warkentin, Ludovic Peran, Minh Giang, Clément Farabet, Oriol Vinyals, Jeff Dean, Koray Kavukcuoglu, Demis Hassabis, Zoubin Ghahramani, Douglas Eck, Joelle Barral, Fernando Pereira, Eli Collins, Armand Joulin, Noah Fiedel, Evan Senter, Alek Andreev, and Kathleen Kenealy.
\newblock Gemma: Open models based on gemini research and technology, 2024.
\newblock URL \url{https://arxiv.org/abs/2403.08295}.

\bibitem[Team(2024{\natexlab{b}})]{qwen1.5}
Qwen Team.
\newblock Introducing qwen1.5, February 2024{\natexlab{b}}.
\newblock URL \url{https://qwenlm.github.io/blog/qwen1.5/}.

\bibitem[Team(2024{\natexlab{c}})]{qwen_moe}
Qwen Team.
\newblock Qwen1.5-moe: Matching 7b model performance with 1/3 activated parameters", February 2024{\natexlab{c}}.
\newblock URL \url{https://qwenlm.github.io/blog/qwen-moe/}.

\bibitem[Teknium(2023)]{OpenHermes}
Teknium.
\newblock Openhermes 2.5: An open dataset of synthetic data for generalist llm assistants, 2023.
\newblock URL \url{https://huggingface.co/datasets/teknium/OpenHermes-2.5}.

\bibitem[Tillet et~al.(2019)Tillet, Kung, and Cox]{10.1145/3315508.3329973}
Philippe Tillet, H.~T. Kung, and David Cox.
\newblock Triton: an intermediate language and compiler for tiled neural network computations.
\newblock In \emph{Proceedings of the 3rd ACM SIGPLAN International Workshop on Machine Learning and Programming Languages}, MAPL 2019, pp.\  10–19, New York, NY, USA, 2019. Association for Computing Machinery.
\newblock ISBN 9781450367196.
\newblock \doi{10.1145/3315508.3329973}.
\newblock URL \url{https://doi.org/10.1145/3315508.3329973}.

\bibitem[Touvron et~al.(2023{\natexlab{a}})Touvron, Lavril, Izacard, Martinet, Lachaux, Lacroix, Rozière, Goyal, Hambro, Azhar, Rodriguez, Joulin, Grave, and Lample]{touvron2023llamaopenefficientfoundation}
Hugo Touvron, Thibaut Lavril, Gautier Izacard, Xavier Martinet, Marie-Anne Lachaux, Timothée Lacroix, Baptiste Rozière, Naman Goyal, Eric Hambro, Faisal Azhar, Aurelien Rodriguez, Armand Joulin, Edouard Grave, and Guillaume Lample.
\newblock Llama: Open and efficient foundation language models, 2023{\natexlab{a}}.
\newblock URL \url{https://arxiv.org/abs/2302.13971}.

\bibitem[Touvron et~al.(2023{\natexlab{b}})Touvron, Martin, Stone, Albert, Almahairi, Babaei, Bashlykov, Batra, Bhargava, Bhosale, Bikel, Blecher, Ferrer, Chen, Cucurull, Esiobu, Fernandes, Fu, Fu, Fuller, Gao, Goswami, Goyal, Hartshorn, Hosseini, Hou, Inan, Kardas, Kerkez, Khabsa, Kloumann, Korenev, Koura, Lachaux, Lavril, Lee, Liskovich, Lu, Mao, Martinet, Mihaylov, Mishra, Molybog, Nie, Poulton, Reizenstein, Rungta, Saladi, Schelten, Silva, Smith, Subramanian, Tan, Tang, Taylor, Williams, Kuan, Xu, Yan, Zarov, Zhang, Fan, Kambadur, Narang, Rodriguez, Stojnic, Edunov, and Scialom]{touvron2023llama2openfoundation}
Hugo Touvron, Louis Martin, Kevin Stone, Peter Albert, Amjad Almahairi, Yasmine Babaei, Nikolay Bashlykov, Soumya Batra, Prajjwal Bhargava, Shruti Bhosale, Dan Bikel, Lukas Blecher, Cristian~Canton Ferrer, Moya Chen, Guillem Cucurull, David Esiobu, Jude Fernandes, Jeremy Fu, Wenyin Fu, Brian Fuller, Cynthia Gao, Vedanuj Goswami, Naman Goyal, Anthony Hartshorn, Saghar Hosseini, Rui Hou, Hakan Inan, Marcin Kardas, Viktor Kerkez, Madian Khabsa, Isabel Kloumann, Artem Korenev, Punit~Singh Koura, Marie-Anne Lachaux, Thibaut Lavril, Jenya Lee, Diana Liskovich, Yinghai Lu, Yuning Mao, Xavier Martinet, Todor Mihaylov, Pushkar Mishra, Igor Molybog, Yixin Nie, Andrew Poulton, Jeremy Reizenstein, Rashi Rungta, Kalyan Saladi, Alan Schelten, Ruan Silva, Eric~Michael Smith, Ranjan Subramanian, Xiaoqing~Ellen Tan, Binh Tang, Ross Taylor, Adina Williams, Jian~Xiang Kuan, Puxin Xu, Zheng Yan, Iliyan Zarov, Yuchen Zhang, Angela Fan, Melanie Kambadur, Sharan Narang, Aurelien Rodriguez, Robert Stojnic, Sergey Edunov, and Thomas
  Scialom.
\newblock Llama 2: Open foundation and fine-tuned chat models, 2023{\natexlab{b}}.
\newblock URL \url{https://arxiv.org/abs/2307.09288}.

\bibitem[Wei et~al.(2023)Wei, Zhao, Zhang, Zhu, Wang, Yang, Li, Cheng, Lü, Hu, Li, Yang, Luo, Wu, Liu, Cheng, Cheng, Zhang, Zhang, Lin, Wang, Ma, Dong, Sun, Chen, Peng, Liang, Yan, Fang, and Zhou]{wei2023skywork}
Tianwen Wei, Liang Zhao, Lichang Zhang, Bo~Zhu, Lijie Wang, Haihua Yang, Biye Li, Cheng Cheng, Weiwei Lü, Rui Hu, Chenxia Li, Liu Yang, Xilin Luo, Xuejie Wu, Lunan Liu, Wenjun Cheng, Peng Cheng, Jianhao Zhang, Xiaoyu Zhang, Lei Lin, Xiaokun Wang, Yutuan Ma, Chuanhai Dong, Yanqi Sun, Yifu Chen, Yongyi Peng, Xiaojuan Liang, Shuicheng Yan, Han Fang, and Yahui Zhou.
\newblock Skywork: A more open bilingual foundation model, 2023.

\bibitem[Wolf et~al.(2020)Wolf, Debut, Sanh, Chaumond, Delangue, Moi, Cistac, Rault, Louf, Funtowicz, Davison, Shleifer, von Platen, Ma, Jernite, Plu, Xu, Le~Scao, Gugger, Drame, Lhoest, and Rush]{wolf-etal-2020-transformers}
Thomas Wolf, Lysandre Debut, Victor Sanh, Julien Chaumond, Clement Delangue, Anthony Moi, Pierric Cistac, Tim Rault, Remi Louf, Morgan Funtowicz, Joe Davison, Sam Shleifer, Patrick von Platen, Clara Ma, Yacine Jernite, Julien Plu, Canwen Xu, Teven Le~Scao, Sylvain Gugger, Mariama Drame, Quentin Lhoest, and Alexander Rush.
\newblock Transformers: State-of-the-art natural language processing.
\newblock In Qun Liu and David Schlangen (eds.), \emph{Proceedings of the 2020 Conference on Empirical Methods in Natural Language Processing: System Demonstrations}, pp.\  38--45, Online, October 2020. Association for Computational Linguistics.
\newblock \doi{10.18653/v1/2020.emnlp-demos.6}.
\newblock URL \url{https://aclanthology.org/2020.emnlp-demos.6/}.

\bibitem[Yang et~al.(2024)Yang, Yang, Zhang, Hui, Zheng, Yu, Li, Liu, Huang, Wei, et~al.]{yang2024qwen2}
An~Yang, Baosong Yang, Beichen Zhang, Binyuan Hui, Bo~Zheng, Bowen Yu, Chengyuan Li, Dayiheng Liu, Fei Huang, Haoran Wei, et~al.
\newblock Qwen2. 5 technical report.
\newblock \emph{arXiv preprint arXiv:2412.15115}, 2024.

\bibitem[Yang(2023)]{Firefly}
Jianxin Yang.
\newblock Firefly (flowing fireflies): Chinese conversational large language model.
\newblock \url{https://github.com/yangjianxin1/Firefly}, 2023.

\bibitem[Yue et~al.(2024)Yue, Zheng, Zhang, and Chen]{yue2024mammoth2}
Xiang Yue, Tuney Zheng, Ge~Zhang, and Wenhu Chen.
\newblock Mammoth2: Scaling instructions from the web.
\newblock \emph{Advances in Neural Information Processing Systems}, 2024.

\bibitem[Zhang \& Sennrich(2019)Zhang and Sennrich]{zhang2019rootmeansquarelayer}
Biao Zhang and Rico Sennrich.
\newblock Root mean square layer normalization, 2019.
\newblock URL \url{https://arxiv.org/abs/1910.07467}.

\bibitem[Zhang et~al.(2024{\natexlab{a}})Zhang, Qu, Liu, Zhang, Lin, Yu, Pan, Cheng, Liu, Lin, Yuan, Zheng, Pang, Du, Liang, Ma, Li, Ma, Lin, Benetos, Yang, Zhou, Ma, Liu, Niu, Wang, Que, Liu, Liu, Guo, Gao, Zhou, Zhang, Zhou, Wang, Bai, Zhang, Zhang, Wang, Yang, Zhao, Zhang, Ouyang, Huang, and Chen]{zhang2024mapneo}
Ge~Zhang, Scott Qu, Jiaheng Liu, Chenchen Zhang, Chenghua Lin, Chou~Leuang Yu, Danny Pan, Esther Cheng, Jie Liu, Qunshu Lin, Raven Yuan, Tuney Zheng, Wei Pang, Xinrun Du, Yiming Liang, Yinghao Ma, Yizhi Li, Ziyang Ma, Bill Lin, Emmanouil Benetos, Huan Yang, Junting Zhou, Kaijing Ma, Minghao Liu, Morry Niu, Noah Wang, Quehry Que, Ruibo Liu, Sine Liu, Shawn Guo, Soren Gao, Wangchunshu Zhou, Xinyue Zhang, Yizhi Zhou, Yubo Wang, Yuelin Bai, Yuhan Zhang, Yuxiang Zhang, Zenith Wang, Zhenzhu Yang, Zijian Zhao, Jiajun Zhang, Wanli Ouyang, Wenhao Huang, and Wenhu Chen.
\newblock Map-neo: Highly capable and transparent bilingual large language model series.
\newblock \emph{arXiv preprint arXiv: 2405.19327}, 2024{\natexlab{a}}.

\bibitem[Zhang et~al.(2024{\natexlab{b}})Zhang, Qu, Liu, Zhang, Lin, Yu, Pan, Cheng, Liu, Lin, et~al.]{zhang2024map}
Ge~Zhang, Scott Qu, Jiaheng Liu, Chenchen Zhang, Chenghua Lin, Chou~Leuang Yu, Danny Pan, Esther Cheng, Jie Liu, Qunshu Lin, et~al.
\newblock Map-neo: Highly capable and transparent bilingual large language model series.
\newblock \emph{arXiv preprint arXiv:2405.19327}, 2024{\natexlab{b}}.

\bibitem[Zhang et~al.(2024{\natexlab{c}})Zhang, Zeng, Wang, and Lu]{zhang2024tinyllamaopensourcesmalllanguage}
Peiyuan Zhang, Guangtao Zeng, Tianduo Wang, and Wei Lu.
\newblock Tinyllama: An open-source small language model, 2024{\natexlab{c}}.
\newblock URL \url{https://arxiv.org/abs/2401.02385}.

\bibitem[Zhao et~al.(2023)Zhao, Gu, Varma, Luo, Huang, Xu, Wright, Shojanazeri, Ott, Shleifer, Desmaison, Balioglu, Damania, Nguyen, Chauhan, Hao, Mathews, and Li]{zhao2023pytorchfsdpexperiencesscaling}
Yanli Zhao, Andrew Gu, Rohan Varma, Liang Luo, Chien-Chin Huang, Min Xu, Less Wright, Hamid Shojanazeri, Myle Ott, Sam Shleifer, Alban Desmaison, Can Balioglu, Pritam Damania, Bernard Nguyen, Geeta Chauhan, Yuchen Hao, Ajit Mathews, and Shen Li.
\newblock Pytorch fsdp: Experiences on scaling fully sharded data parallel, 2023.
\newblock URL \url{https://arxiv.org/abs/2304.11277}.

\bibitem[Zheng et~al.(2025)Zheng, Zhang, Shen, Liu, Lin, Fu, Chen, and Yue]{zheng2025opencodeinterpreterintegratingcodegeneration}
Tianyu Zheng, Ge~Zhang, Tianhao Shen, Xueling Liu, Bill~Yuchen Lin, Jie Fu, Wenhu Chen, and Xiang Yue.
\newblock Opencodeinterpreter: Integrating code generation with execution and refinement, 2025.
\newblock URL \url{https://arxiv.org/abs/2402.14658}.

\bibitem[Zhou et~al.(2023)Zhou, Liu, Xu, Iyer, Sun, Mao, Ma, Efrat, Yu, Yu, Zhang, Ghosh, Lewis, Zettlemoyer, and Levy]{zhou2023limaalignment}
Chunting Zhou, Pengfei Liu, Puxin Xu, Srini Iyer, Jiao Sun, Yuning Mao, Xuezhe Ma, Avia Efrat, Ping Yu, Lili Yu, Susan Zhang, Gargi Ghosh, Mike Lewis, Luke Zettlemoyer, and Omer Levy.
\newblock Lima: Less is more for alignment, 2023.
\newblock URL \url{https://arxiv.org/abs/2305.11206}.

\bibitem[Zoph(2022)]{9835248}
Barret Zoph.
\newblock Designing effective sparse expert models.
\newblock In \emph{2022 IEEE International Parallel and Distributed Processing Symposium Workshops (IPDPSW)}, pp.\  1044--1044, 2022.
\newblock \doi{10.1109/IPDPSW55747.2022.00171}.

\end{thebibliography}
\bibliographystyle{iclr2025_conference}

\clearpage
\appendix

\section{Pretraining Data Detailed Description}
\label{App:pretrain_data}
\renewcommand{\arraystretch}{1.5} 

\begin{table}[!ht]
    \centering
    \begin{tabular}{@{}m{3cm}m{11cm}@{}}
    \toprule
        \textbf{Dataset} & \textbf{Description} \\ 
    \midrule
        \makecell{\href{https://huggingface.co/datasets/Skywork/SkyPile-150B/tree/main/data}{SkyPile-150B} \\ \citep{wei2023skywork}}
         & Consisting of approximately 150 billion tokens and 620 gigabytes of cleaned text data from 233 million web pages, with rigorous filtering and deduplication to ensure quality and mitigate sensitive and biased information. \\ 
    \midrule
        \makecell{\href{https://opendatalab.org.cn/OpenDataLab/WanJuan1_dot_0?source=Q1NETg}{Wanjuan1.0} \\ \citep{he2023wanjuan}} & Composed of processed data from various sources, including web pages, encyclopedias, books, patents, textbooks, and exam questions, with a total volume of data exceeding 500 million documents, amounting to over 1TB (roughly split equally between Chinese and English data) and has undergone meticulous cleaning, deduplication, and value alignment. \\ 
    \midrule
        \makecell{\href{https://huggingface.co/datasets/pleisto/wikipedia-cn-20230720-filtered}{Wikipedia-cn}} & Based on the July 20th, 2023 Chinese Wikipedia dump, retains 254,574 high-quality entries after filtering out special types, low-quality, sensitive, and controversial content, and includes conversions between simplified and traditional Chinese. \\ 
    \midrule    
        \makecell{\href{https://huggingface.co/datasets/xuqinyang/BaiduBaike-5.63M}{Baidu Baike}} & Consisting of 5,630,000 uncleaned entries from Baidu Baike, with a total size of approximately 17GB. \\ 
    \midrule    
        \makecell{\href{https://aistudio.baidu.com/datasetdetail/107726}{Baidu QA}} & Including 1.5 million high-quality encyclopedia questions and answers, spanning 492 categories, with 434 categories occurring at least 10 times, suitable for training intelligent Q\&A systems \\ 
    \midrule
        \makecell{\href{https://huggingface.co/datasets/wangrui6/Zhihu-KOL}{Zhihu QA}} & Including 1 million entries of questions and answers, with 1.5GB in size.\\ 
    \midrule
        \makecell{\href{https://github.com/LianjiaTech/BELLE/tree/main/data/10M}{BELLE} \\ \citep{BELLE}} & Including train\_2M\_CN and train\_3.5M\_CN, which are generated by ChatGPT, containing 2 million and 3.5 million dialogue entries respectively, and both used in this project. Note that these datasets are unverified and may contain errors. \\ 
    \midrule
        \makecell{\href{https://hf-mirror.com/datasets/YeungNLP/moss-003-sft-data}{Moss}\\ \citep{Sun2024MOSS}} & Containing 1.1 million Chinese and English multi-turn dialogue entries. \\ 
    \midrule
        \makecell{\href{https://hf-mirror.com/datasets/YeungNLP/firefly-train-1.1M}{Firefly} \\ \citep{Firefly}} & Comprising 1.15 million entries which cover 23 common Chinese NLP tasks and include culturally relevant data such as couplets, poetry, classical Chinese translations, prose, and Jin Yong's novels, resulting in a total of 1.15 million entries. \\ 
    \midrule
        \makecell{\href{https://hf-mirror.com/datasets/bigcode/starcoderdata}{Starcode} \\ \citep{li2023starcoder} } & Including 783GB of code across 86 programming languages, with 54GB of GitHub Issues, 13GB of Jupyter notebooks, and 32GB of GitHub commits. Our project used only the C++, Python, and Java data. \\ 
    \bottomrule
    \end{tabular}
\caption{Pretraining Data Detailed Description}
\label{table:pretrain_dataset_info}
\end{table}

\clearpage

\section{Supervised Finetuning Data Detailed Description}
\label{App:sft_data}

\begin{table}[!ht]
    \centering
    \begin{tabular}{@{}m{3cm}m{11cm}@{}}
    \toprule
        \textbf{Dataset} & \textbf{Description} \\ 
    \midrule
        \makecell{\href{https://huggingface.co/datasets/BAAI/Infinity-Instruct}{Infinity-Instruct-7M}\\ \citep{InfinityInstruct2024}} & A large-scale, high-quality instruction dataset with only 0.7M of Chinese data used in this project. \\ 
    \midrule
        \makecell{\href{https://zhida.zhihu.com/search/3645185472052319285?zhida_source=entity}{Wanjuan1.0} \\ \citep{he2023wanjuan}} & Consistent with the one used during the pre-training stage, but with the Chinese choice question data repurposed for fine-tuning. \\ 
    \midrule
        \makecell{\href{https://github.com/FunnySaltyFish/Better-Ruozhiba}{Ruozhiba} \\ \citep{better-ruozhiba}} & Questions from Baidu Tieba ``Ruozhiba" were answered by GPT-4, then manually reviewed and edited for formatting errors and improved responses. \\ 
    \midrule
        \makecell{\href{https://github.com/SmartFlowAI/EmoLLM/tree/main}{Self-awareness Dataset}\\ \citep{2024EmoLLM}} & Consisting of various ``Who are you?" questions from the EmoLLM project templates. \\ 
    \midrule
        \makecell{\href{https://huggingface.co/datasets/m-a-p/Code-Feedback}{Code-Feedback} \\ \citep{zheng2025opencodeinterpreterintegratingcodegeneration}} & A code SFT dataset consists of 66,000 entries from various open-source code datasets and LeetCode, after undergoing a series of filtering and selection processes. \\ 
    \midrule
        \makecell{\href{https://huggingface.co/datasets/TIGER-Lab/WebInstructSub}{WebInstructSub} \\ \citep{yue2024mammoth2}} & Containing 2.33 million SFT entries across fields such as mathematics, physics, biology, chemistry, and computer science. \\ 
    \midrule
        \makecell{\href{https://huggingface.co/datasets/teknium/OpenHermes-2.5}{OpenHermes-2.5}\\ \citep{OpenHermes}} & Consisting of samples synthesized by large models and chat data, filtered from open-source data like Airoboros, ChatBot Arena, and Evol Instruct, totaling 1 million entries. \\ 
    \bottomrule
    \end{tabular}
\caption{Supervised Finetuning Data Detailed Description}
\label{table:sft_dataset_info}
\end{table}

\clearpage
\renewcommand{\arraystretch}{1} 

\section{Data Juicer Operators Used for Text Processing}
\label{App:operators}
\begin{table}[!ht]
    \centering
    \begin{tabular}{@{}p{4.8cm}p{7.5cm}p{3cm}@{}}
    \toprule
        \textbf{Operator} & \textbf{Description} & \makecell{\textbf{Note}} \\ 
    \midrule
        \multirow{2}{*}{chinese\_convert\_mapper} & Converts Chinese between Traditional Chinese, Simplified Chinese and Japanese Kanji & Mode: t2s (tradition to simple) \\
    \midrule
        \multirow{1}{*}{clean\_email\_mapper} & Removes email information & \makecell{-} \\ 
    \midrule
        \multirow{2}{*}{clean\_html\_mapper} & Removes HTML tags and returns plain text of all the nodes & \makecell{-} \\ 
    \midrule
        \multirow{1}{*}{clean\_ip\_mapper} & Removes IP addresses & \makecell{-} \\ 
    \midrule
        \multirow{1}{*}{clean\_links\_mapper} & Removes links, such as those starting with http or ftp & \makecell{-} \\ 
    \midrule
        \multirow{2}{*}{clean\_copyright\_mapper} & Removes copyright notice at the beginning of code files (must contain the word copyright) & \makecell{-} \\ 
    \midrule
        \multirow{2}{*}{expand\_macro\_mapper} & Expands macros usually defined at the top of TeX documents & \makecell{-} \\ 
    \midrule
        \multirow{1}{*}{fix\_unicode\_mapper} & Fixes broken Unicodes & \makecell{-} \\ 
    \midrule
        \multirow{2}{*}{punctuation\_normalization\_mapper} & Normalizes various Unicode punctuations to their ASCII equivalents & \makecell{-} \\ 
    \midrule
        \multirow{1}{*}{remove\_repeat\_sentences\_mapper} & Remove repeat sentences in text samples & \makecell{Ignore special char-\\acter and sentences \\ shorter than 2 will\\not be deduplicated} \\ 
    \midrule
        \multirow{1}{*}{remove\_specific\_chars\_mapper} & Removes any user-specified characters or substrings & \makecell{-} \\ 
    \midrule
        \multirow{2}{*}{whitespace\_normalization\_mapper} & Normalizes various Unicode whitespaces to the normal ASCII space (U+0020) & \makecell{-} \\ 
    \midrule
        \multirow{2}{*}{alphanumeric\_filter} & Keeps samples with alphanumeric ratio within the specified range & \makecell{[0.0, 0.9]} \\ 
    \midrule
        \multirow{2}{*}{average\_line\_length\_filter} & Keeps samples with average line length within the specified range & \makecell{[10, 150]} \\ 
    \midrule
        \multirow{2}{*}{character\_repetition\_filter} & Keeps samples with char-level n-gram repetition ratio within the specified range & \makecell{[0.0, 0.4]} \\ 
    \midrule
        \multirow{2}{*}{maximum\_line\_length\_filter} & Keeps samples with maximum line length within the specified range & \makecell{1000} \\ 
    \midrule
        \multirow{2}{*}{perplexity\_filter} & Keeps samples with perplexity score below the specified threshold & \makecell{1500} \\ 
    \midrule
        \multirow{2}{*}{special\_characters\_filter} & Keeps samples with special-char ratio within the specified range & \makecell{[0.0, 0.25]} \\ 
    \midrule
        \multirow{2}{*}{text\_length\_filter} & Keeps samples with total text length within the specified range & \makecell{[10, 100000]} \\ 
    \midrule
        \multirow{2}{*}{word\_repetition\_filter} & Keeps samples with word-level n-gram repetition ratio within the specified range & \makecell{[0.0, 0.5]} \\ 
    \midrule
        \multirow{2}{*}{document\_simhash\_deduplicator} & Deduplicates samples at document-level using SimHash & \makecell{Tokenization:space; \\ window\_size:6; \\ num\_blocks:6; \\ hamming\_distance:4; \\ lowercase:true} \\ 
    \bottomrule
    \end{tabular}
\caption{Data Juicer Operators Used for Text Processing}
\label{table:data_juicer_operators_text}
\end{table}

\clearpage

\section{Data Juicer Operators Used for Code Processing}
\renewcommand{\arraystretch}{1.5} 

\label{App:code_processing}
\begin{table}[!ht]
    \centering
    \begin{tabular}{@{}p{5cm}p{7cm}p{3.3cm}@{}}
    \toprule
        \textbf{Operator} & \textbf{Description} & \makecell{\textbf{Note}} \\ 
    \midrule
        \multirow{2}{*}{\makecell{clean\_copyright\_mapper}} & Removes copyright notice at the beginning of code files (must contain the word copyright) & \makecell{-} \\ 
    \midrule
        clean\_email\_mapper & Removes email information & \makecell{-} \\ 
    \midrule
        \multirow{2}{*}{\makecell{clean\_links\_mapper}} & Removes links, such as those starting with http or ftp & \makecell{-} \\ 
    \midrule
        fix\_unicode\_mapper & Fixes broken Unicodes & \makecell{-} \\ 
    \midrule
        \multirow{2}{*}{\makecell{punctuation\_normalization\_mapper}} & Normalizes various Unicode punctuations to their ASCII equivalents & \makecell{-} \\ 
    \midrule
        \multirow{2}{*}{\makecell{alphanumeric\_filter}} & Keeps samples with alphanumeric ratio within the specified range & \makecell{[0.546, 3.65]} \\ 
    \midrule
        \multirow{2}{*}{\makecell{average\_line\_length\_filter}} & Keeps samples with average line length within the specified range & \makecell{[10, 150]} \\ 
    \midrule
        \multirow{2}{*}{\makecell{character\_repetition\_filter}} & Keeps samples with char-level n-gram repetition ratio within the specified range & \makecell{0.36} \\ 
    \midrule
        \multirow{2}{*}{\makecell{maximum\_line\_length\_filter}} & Keeps samples with maximum line length within the specified range & \makecell{1000} \\ 
    \midrule
        \multirow{2}{*}{\makecell{text\_length\_filter}} & Keeps samples with total text length within the specified range & \makecell{96714} \\ 
    \midrule
        \multirow{2}{*}{\makecell{words\_num\_filter}} & Keeps samples with word count within the specified range & \makecell{[20,6640]} \\ 
    \midrule
        \multirow{2}{*}{\makecell{word\_repetition\_filter}} & Keeps samples with word-level n-gram repetition ratio within the specified range & \makecell{[10, 0.357]} \\ 
    \midrule
        \multirow{2}{*}{\makecell{document\_simhash\_deduplicator}} & Deduplicates samples at document-level using SimHash & \makecell{Tokenization:space; \\ window\_size:6; \\ num\_blocks:6; \\ hamming\_distance:4; \\ lowercase:true} \\ 
    \bottomrule
    \end{tabular}
\caption{Data Juicer Operators Used for Code Processing}
\label{table:data_juicer_operators}
\end{table}

\clearpage

\section{Pre-training Loss Curve}
\label{App:loss_curve}
\begin{figure}[H]
    \centering
    \includegraphics[width=1\textwidth]{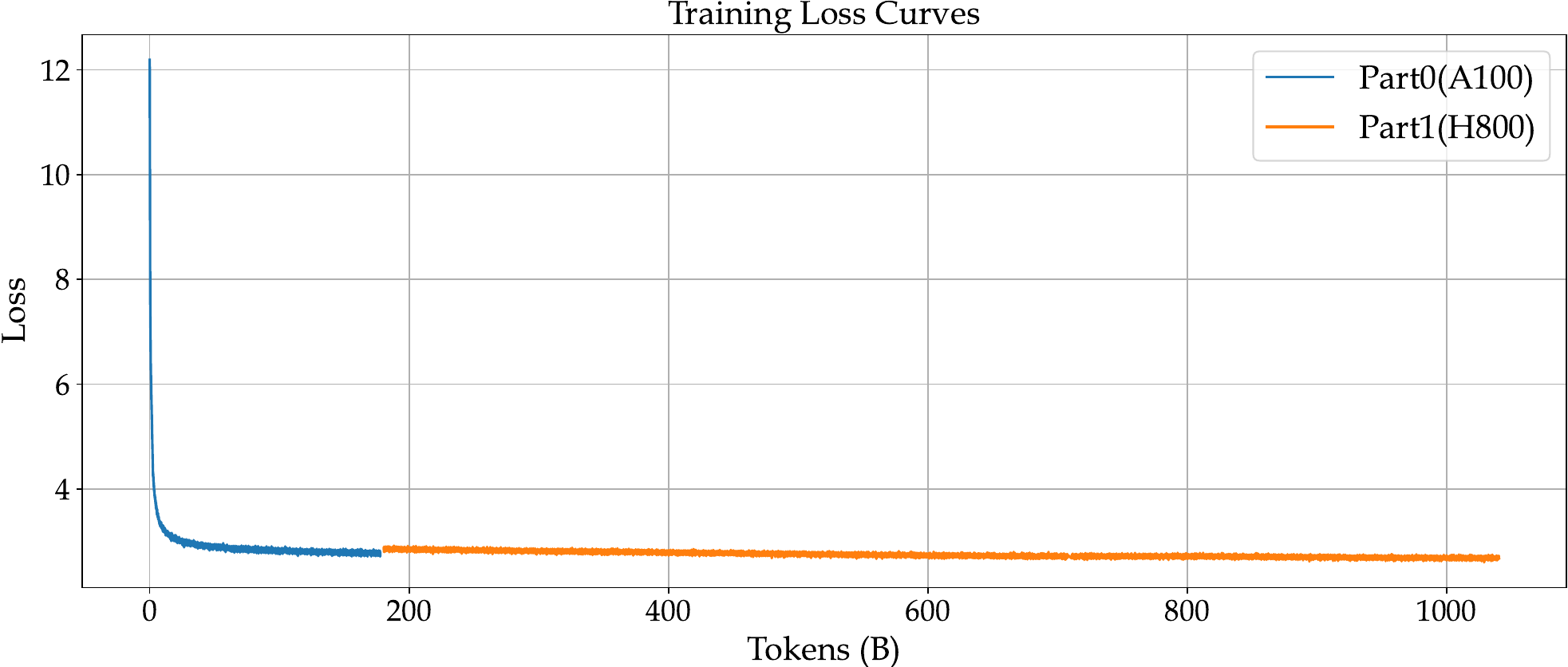}
    \caption{Pre-training loss curve for \model}
    \label{fig:wandb_loss}
\end{figure}

The loss curve during the pre-training stage is shown in \autoref{fig:wandb_loss}.  The initial 200,000 steps utilized NVIDIA 8 A100-80G GPUs for training, while the remaining steps employed 8 H800-80G GPUs.

\section{Supervised Finetuning Loss Curve}
\label{App:sft_loss_curve}

\begin{figure}[H]
    \centering
    \includegraphics[width=1\textwidth]{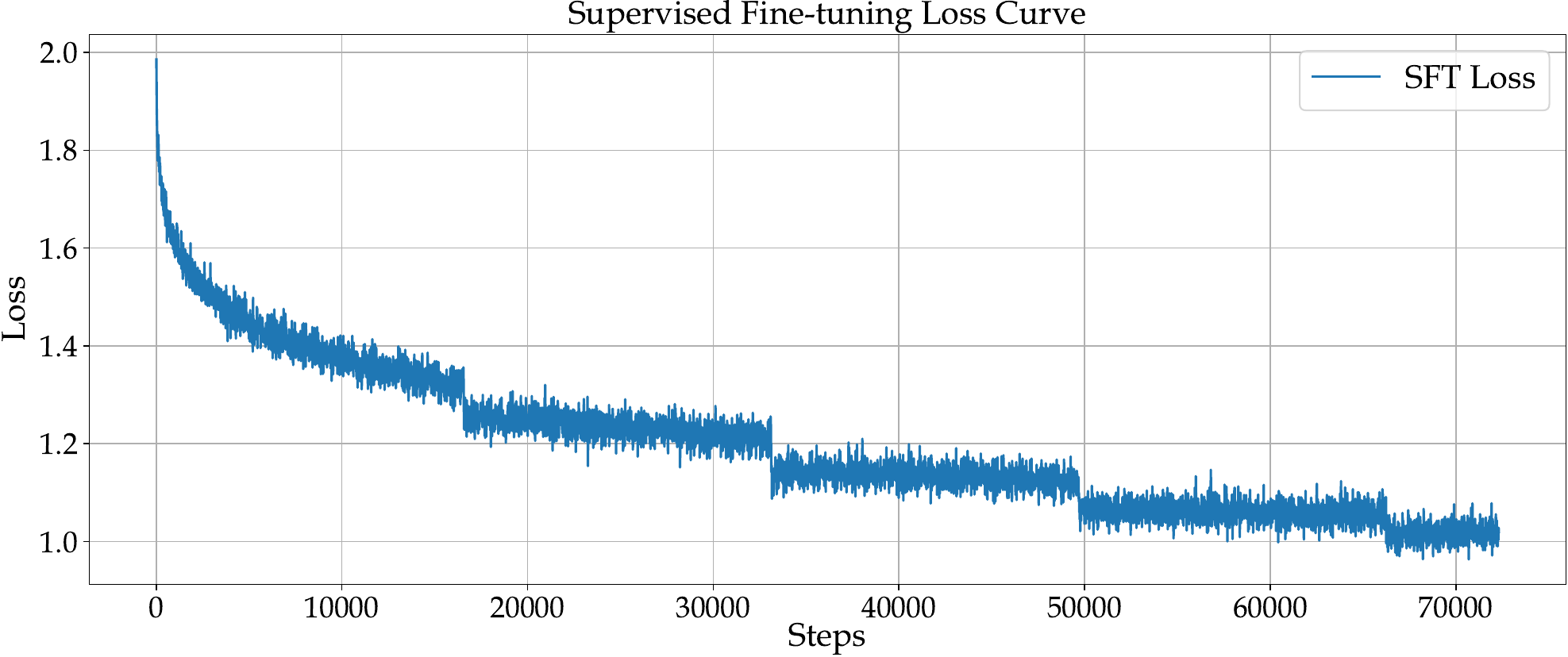}
    \caption{Supervised Fine-tuning loss curve for \model}
    \label{fig:sft_loss}
\end{figure}

\section{Direct Preference Optimization Loss Curve}
\label{App:dpo_loss_curve}

\begin{figure}[H]
    \centering
    \includegraphics[width=1\textwidth]{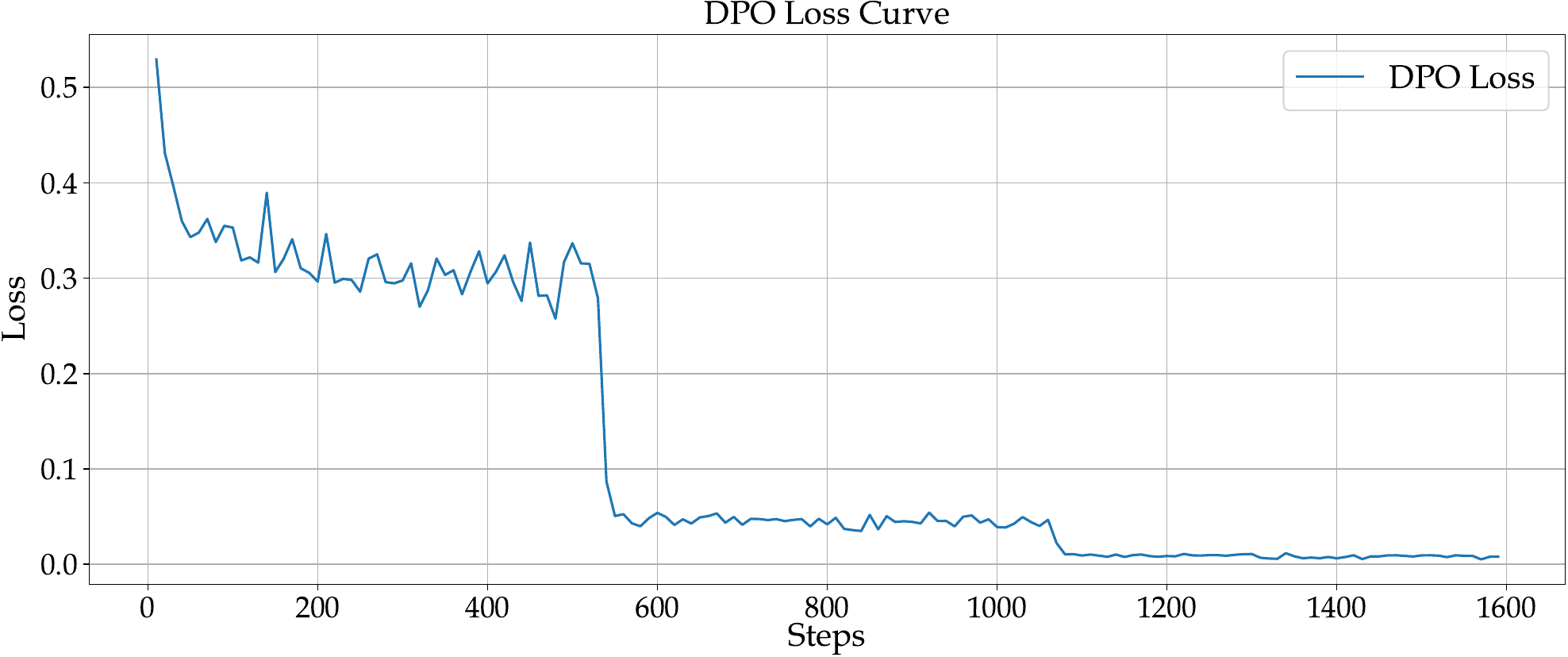}
    \caption{Direct Preference Optimization loss curve for \model}
    \label{fig:dpo_loss}
\end{figure}

\end{document}